\documentclass{article}

\usepackage{microtype}
\usepackage{graphicx}
\usepackage{subcaption}
\usepackage{booktabs} % for professional tables
\usepackage{multirow}

\usepackage{hyperref}

\usepackage[preprint]{icml2026}

\usepackage{amsmath}
\usepackage{amssymb}
\usepackage{mathtools}
\usepackage{amsthm}
\usepackage{listings}
\usepackage{bm}

\usepackage[capitalize,noabbrev]{cleveref}

\theoremstyle{plain}

\theoremstyle{definition}

\theoremstyle{remark}

\usepackage[disable,textsize=tiny]{todonotes}
\icmltitlerunning{\textbf{C}o-\textbf{O}ptimization of \textbf{S}ymbolic \textbf{I}nteractions and \textbf{N}etwork \textbf{E}dges}

\begin{document}

\twocolumn[
  \icmltitle{Interpretable Relational Inference with LLM-Guided Symbolic Dynamics Modeling}



  \icmlsetsymbol{equal}{*}

  \begin{icmlauthorlist}
    \icmlauthor{Xiaoxiao Liang}{equal,ustc}
    \icmlauthor{Juyuan Zhang}{equal,ustc}
    \icmlauthor{Liming Pan}{ustc}
    \icmlauthor{Linyuan L\"u}{ustc}
  \end{icmlauthorlist}
  
  \icmlaffiliation{ustc}{School of Cyber Science and Technology, University of Science and Technology of China, Hefei, Anhui, China}
  
  \icmlcorrespondingauthor{Liming Pan}{pan\_liming@ustc.edu.cn}
  \icmlcorrespondingauthor{Linyuan L\"u}{linyuan.lv@ustc.edu.cn}
  
  \icmlkeywords{Relational Inference, Neuro-symbolic AI, Interpretability, Machine Learning, ICML}

  \vskip 0.3in
]



\printAffiliationsAndNotice{\icmlEqualContribution}

\begin{abstract}
    Inferring latent interaction structures from observed dynamics is a fundamental inverse problem in many-body interacting systems. Most neural approaches rely on black-box surrogates over trainable graphs, achieving accuracy at the expense of mechanistic interpretability. Symbolic regression offers explicit dynamical equations and stronger inductive biases, but typically assumes known topology and a fixed function library. We propose \textbf{COSINE} (\textbf{C}o-\textbf{O}ptimization of \textbf{S}ymbolic \textbf{I}nteractions and \textbf{N}etwork \textbf{E}dges), a differentiable framework that jointly discovers interaction graphs and sparse symbolic dynamics. To overcome the limitations of fixed symbolic libraries, COSINE further incorporates an outer-loop large language model that adaptively prunes and expands the hypothesis space using feedback from the inner optimization loop. Experiments on synthetic systems and large-scale real-world epidemic data demonstrate robust structural recovery and compact, mechanism-aligned dynamical expressions. Code: \url{https://anonymous.4open.science/r/COSINE-6D43}.
\end{abstract}

\section{Introduction}
\label{intro}

Interacting structures among constituent entities are fundamental to understanding dynamical systems across physics, biology, neuroscience, epidemiology, finance, and sociology~\citep{arenas2008synchronization, zhu2022biology, izhikevich2007neurscience, pastor2015epidemic, moraffah2021causalfinance, castellano2009statistical}. In many settings, graphs are unobserved due to experimental, ethical, or economic constraints, while time-resolved dynamics are measurable, motivating inference of latent interactions from trajectories.

\begin{figure}[t]
    \vskip 0.1in
    \begin{center}
    \centerline{\includegraphics[width=0.8\linewidth]{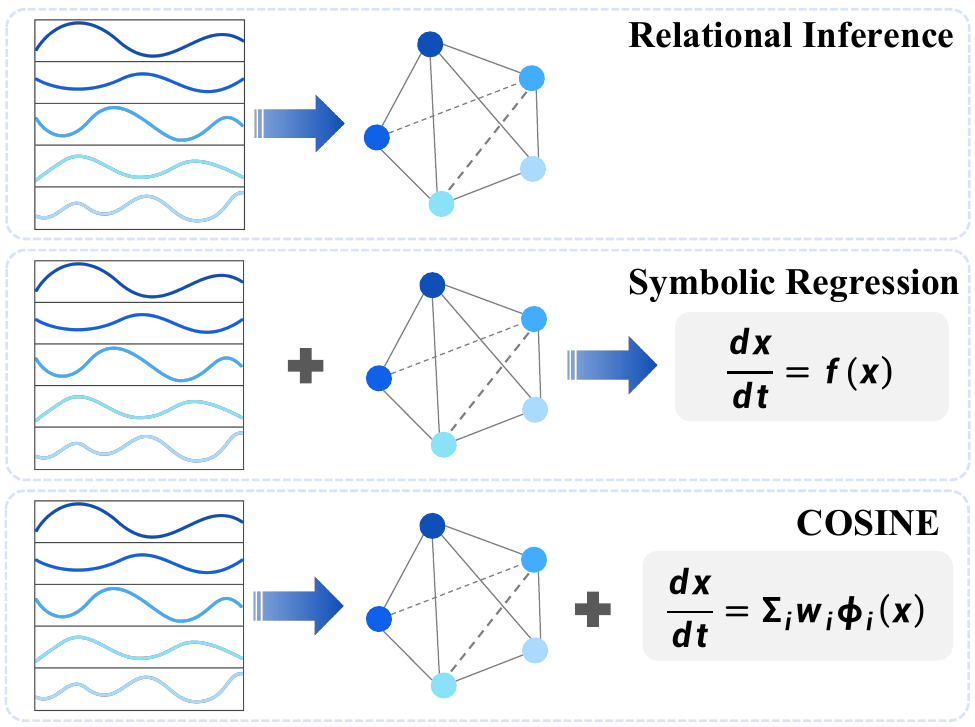}}
    \caption{Conceptual comparison of relational inference, symbolic regression, and COSINE. Relational inference recovers latent interaction graphs from dynamical data; symbolic regression discovers governing equations of dynamics given known structures; COSINE integrates sparse symbolic modeling with graph learning to jointly infer both structure and dynamics.}
    \label{fig:illustration}
    \vspace{-15pt}
    \end{center}
\end{figure}

% As illustrated in \cref{fig:illustration}, relational inference estimates pairwise dependencies from dynamical data. Classical methods, including Granger causality~\citep{granger1969investigating}, mutual information~\citep{wu2020discovering}, and transfer entropy~\citep{schreiber2000measuring}, focus on statistical dependencies but neglect explicit dynamical modeling. Neural approaches, such as Neural Relational Inference (NRI) and its extensions~\citep{kipf2018neural, graber2020dNRI, pan2024gdp, lowe2022amortized, chen2021nrimpm}, jointly infer latent structures and dynamics, yet scale poorly due to edge-centric message passing neural network~\citep{gilmer2017neural}. Attention-based models~\citep{lu2023attention, wu2025riva} improve efficiency via attention but remain limited in modeling local dynamical evolution. Critically, these black-box neural methods exploit over-parameterization to fit spurious edges, hindering mechanistic interpretability.

As illustrated in \cref{fig:illustration}, relational inference estimates pairwise dependencies from dynamical data. Classical methods, including Granger causality~\citep{granger1969investigating}, mutual information~\citep{wu2020discovering}, and transfer entropy~\citep{schreiber2000measuring}, focus on statistical dependencies but neglect explicit dynamical modeling. Neural approaches, such as Neural Relational Inference (NRI) and its extensions~\citep{kipf2018neural, graber2020dNRI, pan2024gdp, lowe2022amortized, chen2021nrimpm}, jointly infer latent structures and dynamics. Attention-based models~\citep{lu2023attention, wu2025riva} improve efficiency via attention, they remain limited by a lack of mechanistic transparency. Critically, these black-box neural methods tend to exploit over-parameterization to fit spurious edges, which accommodates the observed data at the expense of discovering the true governing mechanisms.

% As illustrated in \cref{fig:illustration}, relational inference estimates pairwise dependencies from dynamical data. Traditional statistical approaches, such as Granger causality~\citep{granger1969investigating}, mutual information~\citep{wu2020discovering} and transfer entropy~\citep{schreiber2000measuring}, identify interaction patterns but neglect dynamical modeling. Neural approaches, such as the pioneering Neural Relational Inference (NRI) model and its extensions~\citep{kipf2018neural, graber2020dNRI, pan2024gdp, lowe2022amortized, chen2021nrimpm}, jointly infer latent structures and dynamics, but suffer from poor scalability due to edge-centric message passing neural network~\citep{gilmer2017neural}. Transformer architectures~\citep{lu2023attention, wu2025riva} improve efficiency through self-attention, yet remain limited in precisely modeling local dynamical evolution. Critically, these black-box neural methods exploit over-parameterization to fit spurious edges, hindering mechanistic interpretability.

Symbolic regression (SR)~\cite{schmidt2009distilling} instead targets explicit governing equations from dynamics~\citep{udrescu2020aifeyman, cranmer2019learningSR, cranmer2020discoveringPySR, wong2022automatedpysrWave, kamienny2022transformerSR}. Concise expressions impose strong inductive bias and improve mechanistic readability; expression-level structure is further supported by recent compositional models, including KAN and related studies~\citep{liu2024kan, interpretable_ml_physics_review, makke2024interpretable}. This compactness acts as a functional constraint: incorrect edges are harder to absorb through over-parameterization.

Despite these advantages, symbolic regression on network dynamics remains challenging when networks are unknown. As shown in \cref{fig:illustration}, most existing SR methods assume known network~\citep{grayeli2024symbolicLaSR, yu2025discoverND2, ruan2025discoveringPSE, tsoi2025symbolfit}. With latent structure and dynamics, discovery becomes a coupled search; SINDy reduces equation discovery to sparse regression over fixed libraries~\citep{brunton2016sindy, basiri2024sindyg, hoffmann2019reactivesindy, stankovic2020datasindy, zhao2025physicssindy, goyal2022rk4sindy, lai2021cseqdiscovery}, but closed-world libraries require domain knowledge and can induce spurious edges and degrade robustness.

To address these challenges, we propose \textbf{COSINE} (\textbf{C}o-\textbf{O}ptimization of \textbf{S}ymbolic \textbf{I}nteractions and \textbf{N}etwork \textbf{E}dges), a differentiable framework for joint structure and mechanism discovery (\cref{fig:illustration}). COSINE decomposes dynamics into reusable \emph{message} and \emph{update} components as sparse combinations over shared symbolic libraries, enabling expression reuse and a search space independent of network size. To relax fixed-library bias, COSINE uses an outer-loop large language model (LLM) as a \emph{symbolic hypothesis proposal mechanism} to prune and augment terms from inner-loop feedback, without system-specific templates. Our contributions are summarized as follows:

\begin{itemize}
\item \textbf{COSINE Framework:} We propose an end-to-end differentiable pipeline that jointly optimizes latent graph structures and sparse dynamical equations, enabling the discovery of interpretable dynamical models under unknown networks.
\item \textbf{Sparse Symbolic Message Passing:} We cast the joint discovery problem as sparse regression over message/update libraries. This constraint prevents over-parameterized dynamics from masking spurious edges, thereby enhancing structural identifiability.
\item \textbf{LLM-Guided Library Evolution:} We utilize a feedback-driven pruning/augmentation strategy to adaptively adjust the basis library without hand-crafted, system-specific templates. This approach decouples LLM-driven \emph{hypothesis generation} from optimization-based \emph{hypothesis selection}, ensuring the discovered mechanisms remain numerically grounded.
\item \textbf{Extensive Experiments:} Evaluations across diverse synthetic dynamical systems and real-world datasets demonstrate that COSINE consistently achieves state-of-the-art performance in relational inference while successfully recovering parsimonious and physically consistent symbolic governing laws.
\end{itemize}

% Two design choices make this joint discovery practically identifiable and usable. First, COSINE restricts the dynamics surrogate to a \emph{sparse} symbolic message-passing form, so incorrect edges are less likely to be ``explained away'' by overparameterized dynamics. Second, we decouple \emph{hypothesis generation} from \emph{hypothesis selection}: the LLM only edits the \emph{candidate} library, while all term selection is decided by the inner-loop objective (validation loss and sparsity), ensuring that the final mechanisms remain numerically grounded.

% \begin{itemize}
% \item \textbf{COSINE framework:} Jointly optimizes latent graphs and sparse dynamical equations in an end-to-end differentiable pipeline.
% \item \textbf{Sparse symbolic message passing:} Casts joint discovery as sparse regression over message/update libraries.
% \item \textbf{LLM-guided library evolution:} Uses feedback-driven pruning/augmentation to adapt the basis library without hand-designed, system-specific terms.
% \end{itemize}

\section{Related Work}
\label{related}

\subsection{Relational Inference}

Inferring latent interaction structures from multivariate dynamical data has been widely studied. Classical techniques characterize interaction structures in observational data using statistical measures, including Granger causality~\cite{granger1969investigating}, correlation and partial correlation~\cite{peng2009partial}, mutual information~\cite{wu2020discovering}, and transfer entropy~\cite{schreiber2000measuring}. These approaches focus on statistical dependencies, neglect explicit dynamical modeling, and fail under complex nonlinear dynamics.

% To mitigate structural collapse and improve identifiability, NRI incorporates a KL regularization term that aligns the inferred graph distribution with a prior.

Neural Relational Inference (NRI)~\cite{kipf2018neural} introduced a unified framework that jointly infers latent graphs and system dynamics via end-to-end learning. An encoder encodes graph from observed trajectories, while a decoder predicts future states conditioned on the inferred graph. Subsequent extensions address time-varying graphs~\cite{graber2020dNRI}, heterogeneous interactions~\cite{webb2019fNRI, ha2023heterlearning}, directed graph inference~\cite{wang2022isidg}, and sparsity/degree priors~\cite{li2019structure, chen2021nrimpm}. Related work also integrates attention-based architectures to capture fine-grained, potentially asymmetric dependencies~\cite{ha2020towards, ha2022learning, lu2023attention, wu2025riva}. However, despite their expressive power, these methods largely treat dynamics as a black box, limiting their interpretability and making them susceptible to spurious associations with over-parameterization.

\subsection{Symbolic and Interpretable Dynamical Modeling}

\paragraph{Sparse Symbolic Regression on Dynamics.} Sparse Identification of Nonlinear Dynamics (SINDy)~\cite{brunton2016sindy} identifies governing equations from dynamical systems by selecting a parsimonious subset of basis functions from a predefined library via sparse optimization. Numerous extensions improve its robustness under noise and higher-order integration schemes~\cite{mangan2017sindyhidden, brunton2016sindyC, goyal2022rk4sindy, zhao2025physicssindy}. While network-aware variants like SINDyG~\cite{basiri2024sindyg, gao2022autonomous} exist, they typically \textbf{assume a fixed or known graph structure}, making them difficult to apply when the network is hidden and no domain knowledge is available.

\paragraph{Symbolic Discovery on Network Dynamics.} To explore broader symbolic spaces, neural-guided symbolic regression has emerged, utilizing evolutionary search~\cite{schmidt2009distilling, cranmer2020discoveringPySR}, neural approximations~\cite{udrescu2020aifeyman}, or Transformer-based generation~\cite{kamienny2022transformerSR, biggio2021NeSymRes}. For networked systems, recent methods like ND2~\cite{yu2025discoverND2} and PSE~\cite{ruan2025discoveringPSE} decompose dynamics into node-, interaction-, and aggregation-level operators, while LaSR~\cite{grayeli2024symbolicLaSR} leverages large-model-assisted reasoning to refine hypotheses, introducing an LLM into its genetic algorithm to improve performance.

Neural relational inference methods often lack interpretability and can overfit spurious interactions. Symbolic regression offers stronger inductive biases for dynamical systems, but becomes difficult when the graph is unknown, since the graph structure and equation form are tightly coupled and the search space explodes. SINDy alleviates symbolic search via sparsity, yet relies on a pre-defined operator library and domain priors. \textbf{COSINE} addresses this gap by jointly optimizing a latent interaction graph and sparse symbolic dynamics, enabling accurate structure inference together with interpretable mechanism discovery.

\section{Method}\label{sec:method}

\subsection{Overall Framework}

\begin{figure*}[!t]
    \vskip 0.1in
    \begin{center}
    \centerline{\includegraphics[width=0.98\linewidth]{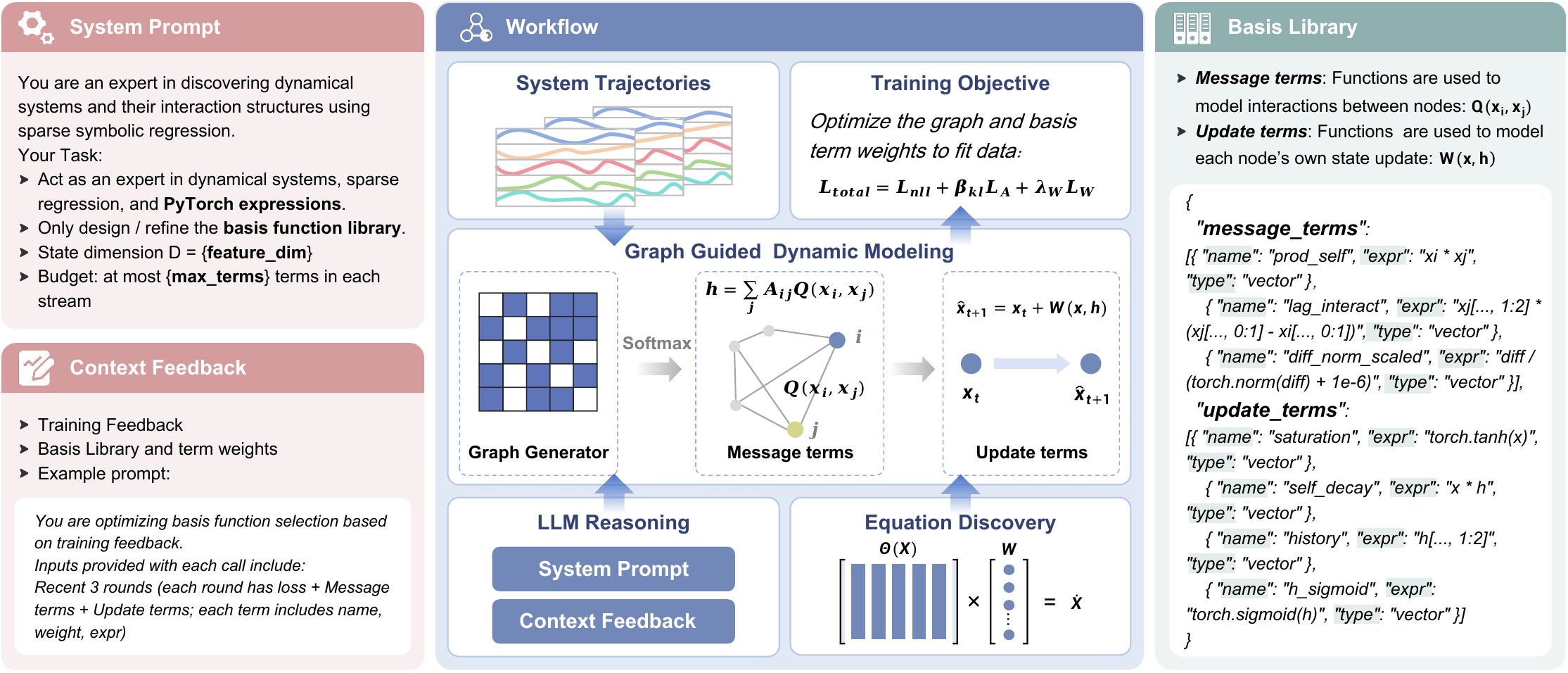}}
    \caption{The COSINE architecture. (Left) LLM-based Reasoning refines the basis library $\Theta(\cdot)$ via performance feedback. (Middle) Graph-Guided Modeling co-optimizes latent edges $A_{ij}$ and coefficients $\mathbf{W}$ through differentiable symbolic message-passing. (Right) Symbolic Basis Library bridges high-level reasoning with numerical discovery of governing mechanisms.}
    \label{fig:framework}
    \end{center}
    \vskip -0.1in
\end{figure*}

COSINE (\textbf{C}o-\textbf{O}ptimization of \textbf{S}ymbolic \textbf{I}nteractions and \textbf{N}etwork \textbf{E}dges) aims to \textbf{jointly discover} the latent interaction graph and governing mechanisms from observed trajectories $\mathbf{X}=(\mathbf{x}^{0},\dots,\mathbf{x}^{T-1})$. We consider $N$ interacting variables, where the state of node $i$ at discretized time $t$ is $x_{i}^{t} \in \mathbb{R}^D$. Let $\mathbf{x}^{t}=\{x_{1}^{t},\dots,x_{N}^{t}\}$ be the global state and $\mathbf{x}_{i}=(x_{i}^{0},\dots,x_{i}^{T-1})$ be the trajectory of node $i$. We write $x_{i}^{t,(d)}$ for the $d$-th component (omitting $t$ in tables as $x_{i}^{(d)}$).

As shown in \cref{fig:framework}, COSINE operates as a two-level closed-loop system. The \textbf{outer loop} employs an LLM as a symbolic supervisor to iteratively refine the basis function library $\Theta(\cdot)$ based on training feedback. Within the \textbf{inner loop}, we adopt a non-amortized setting where the graph $A$ (parameterized by $\bm{\Psi}$) and dynamical coefficients $\mathbf{W}$ are \textbf{co-optimized} via gradient descent to minimize the prediction error. This design enables the discovery of parsimonious, interpretable governing mechanisms without requiring system-specific prior knowledge.

Unless otherwise specified, we train COSINE using one-step supervision (predicting $\mathbf{x}^{t+1}$ from $\mathbf{x}^t$) under teacher forcing, which provides a stable and comparable objective across continuous- and discrete-time systems. At evaluation time, the learned mechanisms can also be rolled out for multiple steps to inspect qualitative trajectory behavior, while the inferred graph is assessed directly by edge-level metrics.

\subsection{Graph Generation Module}

The graph generator encodes unknown interaction structures into a set of trainable latent variables and maps them to an adjacency matrix. For binary relations, we let $a\in\{0,1\}$ denote the edge type. For each ordered pair $(i,j)$, we maintain two real-valued latent variables
$\Psi_{ij}^{0},\Psi_{ij}^{1}\in\mathbb{R}$. To obtain differentiable edge sampling and soft weights, we employ the temperature-controlled Gumbel--Softmax to map latent variables to edge-type probabilities:
\begin{equation}
    A_{ij}^{a}=\mathrm{softmax}\!\Big(\tfrac{1}{\tau}\,[\Psi_{ij}^{0}+g_{ij}^{0},\ \Psi_{ij}^{1}+g_{ij}^{1}]\Big)_{a},
\end{equation}
where $a\in\{0,1\}$ denotes the edge types, $\tau>0$ is the temperature parameter controlling the sharpness of the distribution, and $g_{ij}^{a}$ denotes Gumbel noise to improve graph diversity. As $\tau$ decreases, the edge distribution becomes increasingly peaked and converges to a deterministic graph in the limit. This construction naturally generalizes to multi-relation settings by assigning multi-dimensional latent scores to each edge and applying a Softmax to obtain a categorical distribution over relation types.

In subsequent dynamical modeling, we use the soft adjacency $A_{ij}^{\mathrm{soft}}$ as the weight for message aggregation, enabling differentiable structure learning during training.

\subsection{Sparse Regression Modeling}

Complex networked dynamics can often be written as a combination of \emph{self-dynamics} and \emph{interaction dynamics}:
\begin{equation}
    \frac{d}{dt}x_{i} = f_{i}(\mathbf{x},\mathbf{A}) = W(x_{i}) + \sum_{j=1}^{N}A_{ij}\,Q(x_{i},x_{j}),
\end{equation}
where $W(\cdot)$ describes the intrinsic state evolution of each node, $Q(\cdot,\cdot)$ captures interactions with neighboring nodes, and $A_{ij}\in\{0,1\}$ indicates the network. In classical neural approaches, a surrogate message-passing GNN is leveraged to learn the unknown dynamics from data. A $v \to e$ transformation generates edge features, which are then aggregated back to nodes to update node states ($e \to v$):

\begin{align}\label{gnn}
     & v{\to}e:\quad\mathbf{z}_{(i,j)}^{l}=f_{e}^{l}\big([\mathbf{z}_{i}^{l},\mathbf{z}_{j}^{l},\mathbf{x}_{(i,j)}]\big),    \\
     & e{\to}v:\quad\mathbf{z}_{j}^{l+1}=f_{v}^{l}\big([\sum_{i\in\mathcal{N}_j}\mathbf{z}_{(i,j)}^{l},\mathbf{x}_{j}]\big),
\end{align}

where $\mathbf{z}_{i}^{l}$ denotes the embedding of node $v_{i}$ at layer $l$, $\mathbf{z}_{(i,j)}^{l}$ denotes the embedding of edge $e_{(i,j)}$, and $\mathbf{x}_{i}$ and $\mathbf{x}_{(i,j)}$ represent the initial node and edge features, respectively. $\mathcal{N}_{j}$ denotes the set of neighbors with incoming edges to node $j$, and $[\cdot,\cdot]$ indicates vector concatenation. The functions $f_{v}$ and $f_{e}$ are node- and edge-specific neural networks, respectively. The final node embeddings $\mathbf{z}^{L}$ are then used to predict the next-step state of the dynamical system.

Inspired by the message-passing decomposition in \cref{gnn}, we design a sparse regression module that employs \emph{two basis-function libraries} to approximate the two stages of GNN modeling. Specifically, we decompose the fitting process into a \textbf{message flow} and an \textbf{update flow}. The message flow generates interaction messages from neighbors, while the update flow updates node states based on aggregated messages. By modeling each component using a sparse linear combination of basis functions, we can explicitly capture key mechanisms and interpret the equation structure through coefficient sparsity:
\begin{align}
    \mathbf{e}_{ij} & = \text{MESSAGE}(x_{i}, x_{j}),                                                   \\
    x_{i}^{t+1}          & = \text{UPDATE}\big(x_{i}^{t}, \text{AGG}_{j \in \mathcal{N}_{\mathrm{in}}(i)}\mathbf{e}_{ij}^{t}\big).
\end{align}
Here $\mathcal{N}_{\mathrm{in}}(i)$ denotes the incoming neighbors of node $i$ under the adjacency convention above (i.e., indices $j$ such that $A_{ij}=1$).

\paragraph{Message term.}
Given a message basis function set $\Theta_{\mathrm{msg}}=\{\phi_{1},\dots,\phi_{M}\}$, for each directed pair $(i,j)$ we compute
\begin{equation}
    \phi_{m}^{\mathrm{msg}}(x_{i}^{t},x_{j}^{t})\in\mathbb{R}^{D},\quad m=1,\dots,M,
\end{equation}
where $D$ is the node feature dimension. A linear combination with learnable weights yields the edge message vector
\begin{equation}
    \mathbf{e}_{ij}^{t}=\sum_{m=1}^{M}\mathbf{W}_{\mathrm{msg}}[m]\odot \phi_{m}^{\mathrm{msg}}(x_{i}^{t},x_{j}^{t})\in\mathbb{R}^{D},
\end{equation}
where $\odot$ denotes element-wise multiplication and $\mathbf{W}_{\mathrm{msg}}[m]\in\mathbb{R}^{D}$ is the coefficient vector for the $m$-th basis. Given the soft structure $\mathbf{A}^{\mathrm{soft}}$, messages are aggregated for node $i$ as
\begin{equation}
    \mathbf{h}_{i}^{t}=\sum_{j=1}^{N}A_{ij}^{\mathrm{soft}}\,\mathbf{e}_{ij}^{t}.
\end{equation}
% \lm{talk about when LLM enters}
\paragraph{Update term.}
Given an update basis set $\Theta_{\mathrm{upd}}=\{\phi_{1}^{\mathrm{upd}},\dots,\phi_{U}^{\mathrm{upd}}\}$, we construct update features for node $i$ using the current state $x_i^t$, the aggregated message $\mathbf{h}_{i}^{t}$, and the node degree $k_i = \sum_{j} A_{ij}^{\mathrm{soft}}$:
\begin{equation}
    \phi_{n}^{\mathrm{upd}}(x_{i}^{t},\mathbf{h}_{i}^{t}, k_i)\in\mathbb{R}^{D},\quad n=1,\dots,U,
\end{equation}
and obtain the final increment
\begin{equation}
    \Delta \mathbf{x}_{i}^{t}=\sum_{n=1}^{U}\mathbf{W}_{\mathrm{upd}}[n]\odot \phi_{n}^{\mathrm{upd}}(x_{i}^{t},\mathbf{h}_{i}^{t}, k_i).
\end{equation}
Thus, the one-step prediction is
\begin{equation}
    \hat{\mathbf{x}}_{i}^{t+1}=\mathbf{x}_{i}^{t}+\Delta \mathbf{x}_{i}^{t}.
\end{equation}
For continuous-time systems, this can be viewed as an explicit Euler discretization where $\Delta \mathbf{x}_{i}^{t}$ approximates $\Delta t\,\dot{\mathbf{x}}_{i}^{t}$ (with $\Delta t$ absorbed into the learned coefficients). For discrete-time systems, $\Delta \mathbf{x}_{i}^{t}$ acts as a residual update that directly models the transition $\mathbf{x}_{i}^{t}\mapsto \mathbf{x}_{i}^{t+1}$.

Joint optimization manifests at two levels: (i) the structural layer, where $A_{ij}^{\mathrm{soft}}$ represents the probability or strength of each edge; and (ii) the equation layer, where the \textbf{sparsity} of $\mathbf{W}_{\mathrm{msg}}$ and $\mathbf{W}_{\mathrm{upd}}$ determines which basis functions truly participate in generating the dynamics.

\subsection{Training Objective and Loss}

Given the one-step prediction $\hat{\mathbf{x}}^{t+1}$ and the ground-truth observation $\mathbf{x}^{t+1}$, we employ the Gaussian negative log-likelihood (NLL) as the primary objective, which is equivalent to minimizing the scaled squared error under a fixed variance $\sigma^{2}$. The prediction loss is formulated as:
\begin{equation}
    \mathcal{L}_{\mathrm{nll}}=\frac{1}{2\sigma^{2}}\sum_{i=1}^{N}\left\|\hat{\mathbf{x}}_{i}^{t+1}-\mathbf{x}_{i}^{t+1}\right\|_{2}^{2}.
\end{equation}

To stabilize structure learning and impose a prior over edge-type distributions, we include a KL regularization term (computed over all ordered node pairs):
\begin{equation}
    \mathcal{L}_{A}=\mathrm{KL}\big(q(\mathbf{A})\ \|\ \mathrm{r}\big),
\end{equation}
where $q(\mathbf{A})$ denotes the edge distribution induced by the graph generator, and $\mathrm{r}$ is a simple factorized prior over edge types (e.g., a Bernoulli prior favoring sparsity when appropriate; if no prior knowledge is available, we default to a uniform prior). To encourage sparsity in symbolic regression, we apply $\ell_1$ regularization to both sets of coefficients:
\begin{equation}
    \mathcal{L}_{W}=\|\mathbf{W}_{\mathrm{msg}}\|_{1}+\|\mathbf{W}_{\mathrm{upd}}\|_{1}.
\end{equation}
The total loss is
\begin{equation}
    \mathcal{L}_{\mathrm{total}}=\mathcal{L}_{\mathrm{nll}}+\beta_{\mathrm{KL}}\mathcal{L}_{A}+\lambda_{W}\mathcal{L}_{W},
\end{equation}
where $\beta_{\mathrm{KL}}$ and $\lambda_{W}$ are hyperparameters. The full training process is shown in \cref{appd:train_process}. After each inner-loop optimization, we summarize the current library $\Theta$ performance together with the learned coefficient weights and feed this information to the LLM to guide an \emph{evolution} of the basis library for the next inner-loop.

\subsection{LLM-Guided Library Evolution}
\label{sec:llm_library}

The outer loop performs a lightweight search over symbolic space by \emph{editing the basis library}. Each round uses inner-loop feedback (loss, sparsity, and residual patterns) to prompt the LLM to prune or augment terms, producing a candidate library. To avoid expensive multi-branch search while keeping evolution controlled, we use a \textbf{best-so-far} strategy that maintains a single optimal library and only accepts candidates that strictly improve prediction loss.

Formally, we track the best library across rounds:
\begin{equation}
\Theta^{\star}=\arg\min_{\Theta\in\{\Theta^{(0)},\dots,\Theta^{(r)}\}}\mathcal{L}_{\mathrm{val}}(\Theta),
\end{equation}
and update $\Theta^{\star}\leftarrow\widetilde{\Theta}$ only if $\mathcal{L}_{\mathrm{val}}(\widetilde{\Theta})<\mathcal{L}_{\mathrm{val}}(\Theta^{\star})$. This strategy keeps computation predictable and reduces the risk of degradation from noisy edits, while leveraging strong numerical feedback to steer the symbolic space toward parsimonious, physically consistent mechanisms.
In practice, the validation gate decouples symbolic exploration from numerical selection: the LLM only proposes candidates, while the inner-loop loss determines acceptance. If no candidate improves $\mathcal{L}_{\mathrm{val}}$ over successive rounds, the search naturally terminates without accumulating regressions. In \cref{appd:basis_search} \cref{appd:llm_details}, details of the library evolution and LLM (prompts, model) are presented.

\section{Experiments}
\label{sec:experiments}

\subsection{Results on Synthetic Data}

\begin{table*}[!ht]
\centering
% \vspace{-0.2cm}
\begin{minipage}{\textwidth}
\caption{Relational inference performance (AUC \%) on different graphs with $N=50$ nodes across various dynamical systems. In the VOLUME column, a×b corresponds to trajectories × sampled steps. Boldface marks the highest accuracy.}
\label{tab:unweighted_performance} \vskip 0.1in
\begin{center}
\begin{small}
\begin{sc}
\setlength{\tabcolsep}{6pt}
\renewcommand{\arraystretch}{0.85}
\begin{tabular}{llcccccccc}
\toprule \textbf{Dyn.} & \textbf{Graph} & \textbf{Volume} & \textbf{GC} & \textbf{MI} & \textbf{TE} & \textbf{NRI} & \textbf{GDP} & \textbf{RIVA} & \textbf{COSINE}   \\
\midrule \multirow{3}{*}{MM} & ER-50          & $50 \times 10$  & 60.24       & 77.84       & 53.66       & 96.25$\pm$2.22        & 98.31$\pm$1.41        & 52.17$\pm$1.15         & \textbf{99.63$\pm$0.12}  \\
                           & BA-50          & $50 \times 10$  & 62.79       & 88.04       & 63.96       & 82.67$\pm$1.81        & 93.02$\pm$3.94        & 52.83$\pm$1.22         & \textbf{98.07$\pm$0.45}  \\
                           & WS-50          & $50 \times 10$  & 80.33       & 97.77       & 57.56       & \textbf{99.83$\pm$0.15} & 56.77$\pm$2.55        & 53.65$\pm$1.40         & 99.10$\pm$0.25           \\
\midrule \multirow{3}{*}{Diff}       & ER-50          & $50 \times 10$  & 63.84       & 56.00       & 57.63       & 91.87$\pm$7.80        & 93.44$\pm$4.87        & 53.02$\pm$1.35         & \textbf{99.36$\pm$0.10}  \\
                           & BA-50          & $50 \times 10$  & 75.54       & 72.06       & 61.71       & 94.16$\pm$11.62        & 94.41$\pm$3.23        & 52.89$\pm$1.18         & \textbf{96.55$\pm$1.05}  \\
                           & WS-50          & $50 \times 10$  & 73.63       & 76.39       & 90.04       & 99.31$\pm$0.55        & 56.68$\pm$3.12        & 53.72$\pm$1.25         & \textbf{100.00$\pm$0.00} \\
\midrule \multirow{3}{*}{Spr}        & ER-50          & $15 \times 10$  & 50.61       & 72.24       & 76.05       & 99.84$\pm$0.47        & 99.99$\pm$0.02        & 52.82$\pm$1.50         & \textbf{100.00$\pm$0.00} \\
                           & BA-50          & $15 \times 10$  & 54.53       & 91.16       & 84.67       & 98.17$\pm$5.40        & 99.88$\pm$0.36        & 54.16$\pm$1.62         & \textbf{100.00$\pm$0.00} \\
                           & WS-50          & $15 \times 10$  & 55.44       & 50.40       & 51.83       & 96.77$\pm$1.25        & 99.94$\pm$0.05        & 55.20$\pm$1.45         & \textbf{100.00$\pm$0.00} \\
\midrule \multirow{3}{*}{Kura}       & ER-50          & $30 \times 30$  & 54.94       & 64.69       & 64.76       & 82.09$\pm$19.14        & 94.93$\pm$12.94        & 75.98$\pm$2.34         & \textbf{99.99$\pm$0.01}  \\
                           & BA-50          & $30 \times 30$  & 51.50       & 55.46       & 61.87       & 69.70$\pm$18.16        & 90.13$\pm$12.38        & 76.06$\pm$2.15         & \textbf{99.85$\pm$0.14}  \\
                           & WS-50          & $30 \times 30$  & 59.27       & \textbf{100.00} & 96.63    & 50.17$\pm$1.05        & \textbf{100.00$\pm$0.00} & 98.19$\pm$0.55      & \textbf{100.00$\pm$0.00} \\
\midrule \multirow{3}{*}{FJ}         & ER-50          & $20 \times 10$  & 93.00       & 53.66       & 83.64       & 97.67$\pm$1.06        & 99.82$\pm$0.47        & 52.55$\pm$1.10         & \textbf{100.00$\pm$0.00} \\
                           & BA-50          & $20 \times 10$  & 92.87       & 52.32       & 86.88       & 91.62$\pm$4.67        & 92.63$\pm$13.46        & 52.64$\pm$1.32         & \textbf{99.98$\pm$0.02}  \\
                           & WS-50          & $20 \times 10$  & 99.81       & 66.14       & 95.37       & 68.76$\pm$5.25        & 89.51$\pm$3.44        & 52.89$\pm$1.45         & \textbf{100.00$\pm$0.00} \\
\midrule \multirow{3}{*}{CMN}        & ER-50          & $20 \times 10$  & 52.89       & 87.39       & 64.35       & 89.76$\pm$2.59        & 97.58$\pm$3.38        & 50.84$\pm$0.95         & \textbf{99.75$\pm$0.15}  \\
                           & BA-50          & $20 \times 10$  & 53.44       & 87.84       & 71.51       & 83.35$\pm$2.30        & 88.83$\pm$6.19        & 53.87$\pm$1.67         & \textbf{93.23$\pm$2.45}  \\
                           & WS-50          & $20 \times 10$  & 61.08       & 98.94 & 86.70       & 69.21$\pm$4.12        & 51.51$\pm$1.10        & 55.96$\pm$1.88         & \textbf{99.05$\pm$0.55}           \\
\bottomrule
\end{tabular}
\end{sc}
\end{small}
\end{center}
\end{minipage}
\vspace{0.4cm} % 控制两个表之间的距离

\begin{minipage}{\textwidth}
\caption{Dominant terms discovered by COSINE on ER graphs (\cref{tab:unweighted_performance}), ranked by absolute coefficient magnitude. \emph{Term~1--3} denote the top three terms for \emph{Message} and \emph{Update} modules. \emph{Term Acc.} reports row-wise primitive coverage scores. $x$: node state; $h_i$: aggregated message; $k_i$: degree; $\epsilon=10^{-6}$.}
\label{tab:discovered_terms} \vskip 0.1in
\begin{center}
\begin{small}
\begin{sc}
\setlength{\tabcolsep}{4pt}
\renewcommand{\arraystretch}{0.85}
\resizebox{\textwidth}{!}{%
\begin{tabular}{l c c c c c c}
\toprule \textbf{Dyn.} & \textbf{Ground-truth Dynamics }                                                                & \textbf{Component} & \textbf{Term 1}                             & \textbf{Term 2}                         & \textbf{Term 3}                       & \textbf{Term Acc.} \\
                    \midrule \multirow{2}{*}{MM} & \multirow{2}{*}{$\dot x_{i} = -x_{i} + \sum_{j\in\mathcal{N}(i)}\frac{A_{ij}}{k_{i}}\,\frac{x_{j}}{1+x_{j}}$}                 & Message            & $(x_{j} - x_{i})^{2}$ (0.31)                & $|x_{j} - x_{i}|$ (0.28)                & $x_{i} x_{j}$ (0.20)                  & 0.00               \\
                           &                                                                                                              & Update             & $\frac{h_{i}}{k_{i}+\epsilon}$ (0.70)       & $x$ (0.42)                              & $\sin(x)$ (0.25)                      & 1.00               \\
\midrule \multirow{2}{*}{Diff}         & \multirow{2}{*}{$\dot x = -\beta L x,\quad L=D_{\text{in}}-A$}                                                  & Message            & $x_{j} - x_{i}$ (0.17)                      & $x_{i} x_{j}$ (0.12)                    & $\cos(x_{j} - x_{i})$ (0.11)          & 1.00               \\
                           &                                                                                                              & Update             & $\frac{x + h_{i}}{k_{i}+1+\epsilon}$ (0.31) & $\frac{h_{i}}{k_{i}+1+\epsilon}$ (0.30) & $\cos(x)$ (0.08)                      & 1.00               \\
\midrule \multirow{2}{*}{Spr}            & \multirow{2}{*}{$\ddot{\mathbf{p}}_{i} = -k\sum_{j\in\mathcal{N}(i)}A_{ij}(\mathbf{p}_{i}-\mathbf{p}_{j})$}                   & Message            & $x_{j} - x_{i}$ (0.02)                      & $x_{i}^{(0)}x_{j}^{(1)}$ (0.01)         & $\sin(x_{i}^{(0)}x_{j}^{(1)})$ (0.01) & 1.00               \\
                           &                                                                                                              & Update             & $\frac{h_{i}}{k_{i}+1}$ (0.04)              & $\frac{x}{k_{i}+1}$ (0.03)              & $x$ (0.02)                            & 1.00               \\
\midrule \multirow{2}{*}{Kura}          & \multirow{2}{*}{$\dot\phi_{i} = \omega_{i} + \sum_{j\in\mathcal{N}(i)}A_{ij}\kappa\sin(\phi_{j}-\phi_{i})$}                   & Message            & $\sin(x_{j} - x_{i})$ (0.53)                & $|x_{j} - x_{i}|$ (0.16)                & $x_{i}^{(1)}x_{j}^{(2)}$ (0.14)       & 1.00               \\
                           &                                                                                                              & Update             & $\frac{h_{i}}{k_{i}+1}$ (0.48)              & $\frac{x h_{i}}{k_{i}+1}$ (0.19)        & $h_{i}$ (0.08)                        & 1.00               \\
\midrule \multirow{2}{*}{FJ}                & \multirow{2}{*}{$x_{i}^{t+1}= \frac{1}{k_{i}+1}\!\left(\sum_{j\in\mathcal{N}(i)}A_{ij}x_{j}^{t} + s_{i}\right)$}              & Message            & $x_{j} - x_{i}$ (0.25)                      & $x_{i}^{(0)}x_{j}^{(1)}$ (0.03)         & $x_{i} x_{j}$ (0.03)                  & 1.00               \\
                           &                                                                                                              & Update             & $\frac{h_{i}}{k_{i}+1}$ (0.97)              & $x$ (0.06)                              & $h_{i}$ (0.06)                        & 1.00               \\
\midrule \multirow{2}{*}{CMN}               & \multirow{2}{*}{$\theta_{i}^{t+1}= (1-s)f(\theta_{i}^{t}) + \frac{s}{k_{i}}\sum_{j\in\mathcal{N}(i)}A_{ij}f(\theta_{j}^{t})$} & Message            & $x_{i} x_{j}$ (0.33)                        & $(x_{j} - x_{i})^{2}$ (0.29)            & $x_{j} - x_{i}$ (0.12)                & 0.50               \\
                           &                                                                                                              & Update             & $\frac{h_{i}}{k_{i}+1}$ (2.66)              & $\tanh(h_{i})$ (1.18)                   & $x$ (1.12)                            & 1.00               \\
\bottomrule
\end{tabular}%
}
\end{sc}
\end{small}
\end{center}
\end{minipage}

\end{table*}

\paragraph{Networks.}
We conduct experiments on three standard graph families: Erdős--Rényi (ER), Barabási--Albert (BA), and Watts--Strogatz (WS). For ER graphs, the edge probability is set to $p=0.1$; for BA graphs, each new node attaches to $m=2$ existing nodes; for WS graphs, we start from a ring lattice with $k=2$ nearest neighbors per node and rewire edges with probability $p=0.1$.

\paragraph{Dynamical systems.}
We evaluate COSINE on six representative systems covering biochemical reactions, diffusion, mechanical coupling, synchronization, opinion dynamics, and chaotic maps: (1) \emph{Michaelis-Menten kinetics} (MM) \citep{karlebach2008modelling}; (2) \emph{Diffusion} (Diff); (3) \emph{Network of springs} (Spr); (4) \emph{Kuramoto model} (Kura) \citep{kuramoto1975self}; (5) \emph{Friedkin-Johnsen dynamics} (FJ) \citep{friedkin1990social,abebe2018opinion}; and (6) \emph{Coupled map network} (CMN) \citep{garcia2002coupled}. These systems, generated on ER, BA, and WS networks, span continuous and discrete time with diverse nonlinear interactions.

\paragraph{Evaluation metrics.}
We report AUC as our primary metric for relational inference, computed by comparing the predicted edge scores $A_{ij}^{\mathrm{soft}}$ against the binary ground-truth adjacency. In addition, for mechanism discovery, we compute \emph{term accuracy} as primitive coverage: for each module (Message/Update), we take the top $K=3$ terms by absolute coefficient magnitude and measure the fraction of required ground-truth primitives recovered among them (using simple equivalence rules, e.g., matching sinusoidal coupling on $x_j-x_i$ up to sign/constant scaling).

\paragraph{Baselines.}
We compare COSINE with various baselines from three categories: classical statistical methods (GC~\cite{granger1969investigating}, MI~\cite{butte1999mutual}, and TE~\cite{schreiber2000measuring}), neural relational inference models (NRI~\cite{kipf2018neural} and GDP~\cite{pan2024gdp}), and attention-based models (RIVA~\cite{wu2025riva}). We adapt RIVA to the unweighted setting and tune hyperparameters for all baselines using their public implementations to ensure fair comparison in experiments.

\paragraph{Implementation details.}
For COSINE, the initial basis library is generated by the LLM without system-specific templates. We set the Gumbel--Softmax temperature to $\tau=0.3$, $\beta_{\mathrm{KL}}=0.1$, and $\lambda_W=0.1$, and optimize with Adam (lr $0.1$ for structure and $0.005$ for coefficients). Each outer round runs $1000$ inner epochs for up to $10$ rounds, using validation loss and coefficient statistics to refine the library under a controlled edit budget and a maximum library size. We use GPT-OSS (20B) as the default LLM. More details of synthetic experiments are presented in \cref{appd:syn_exp}.

\paragraph{Relational Inference.} 
\cref{tab:unweighted_performance} shows that COSINE achieves state-of-the-art or highly competitive performance, significantly outperforming statistical methods (GC, MI, TE) and matching or surpassing deep learning models (NRI, GDP, RIVA). While statistical approaches struggle with high-order nonlinearities, COSINE's joint optimization of structure and symbolic dynamics ensures robust reconstruction across diverse regimes. Notably, in complex scenarios like Kuramoto system on BA graphs, COSINE maintains a near-perfect AUC (99.85), substantially exceeding NRI and GDP. Although performance margins can be influenced by specific graphs, in practice, across settings, COSINE consistently demonstrates superior effectiveness in disentangling interactions within heterogeneous networks. 
% Moreover, the gains consistently observed here suggests that COSINE adapts across regimes.
% This trend indicates that the sparse symbolic constraints reduce false positives in edge inference, especially when dynamics are strongly nonlinear. 

\paragraph{Mechanism Discovery.} 
As shown in \cref{tab:discovered_terms}, COSINE consistently recovers mechanism-aligned primitives, with a clear strength in identifying aggregation and normalization patterns in the Update module. COSINE captures intended interaction primitives such as $x_j-x_i$ and sinusoidal coupling, while the Message module occasionally includes predictive proxy nonlinearities when multiple bases can approximate similar interaction effects under finite data and noise. Overall, the alignment between inferred terms and ground-truth mechanisms indicates that COSINE goes beyond pure statistical association toward interpretable mechanism discovery; its goal is \emph{interpretable joint discovery} of both latent structure and dynamics, providing sparse, human-readable interaction primitives and aggregation forms.

\begin{figure}[!t]
\vskip 0.1in
\begin{center}
\centerline{\includegraphics[width=\linewidth]{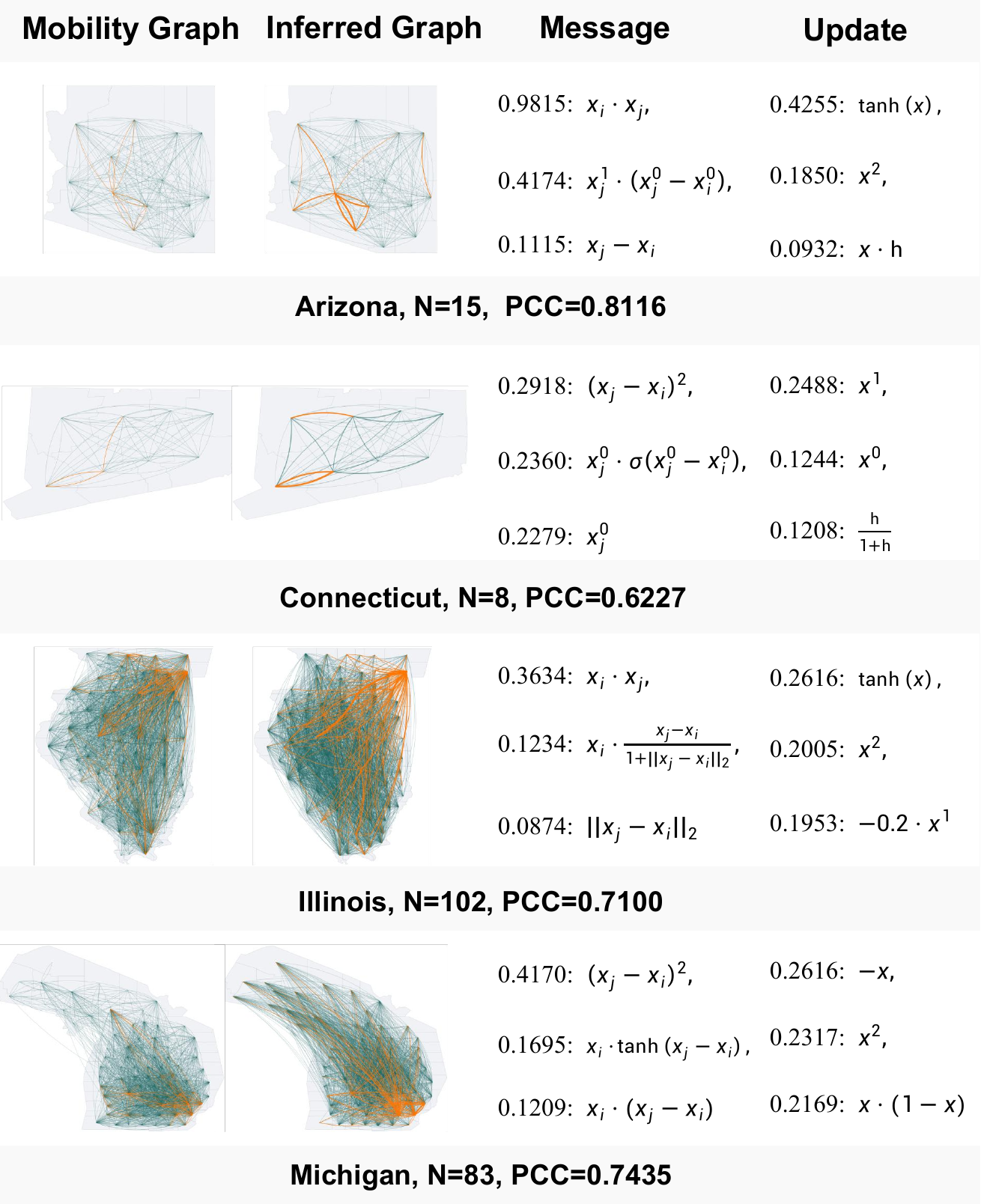}}
\caption{Results on COVID-19 epidemic data. Left: Mobility graph and inferred graph. Right: Learned weights of dominant terms in the Message and Update module across four U.S. states. }
\label{fig:epi_results}
\end{center}
\vspace{-15pt}
\end{figure}

% The framework exhibits consistent role separation: the message module captures pairwise interactions, while the update module recovers aggregation mechanisms, identifying degree-normalized terms ($\frac{h_i}{k_i+1}$) consistent with ground-truth dynamics. Furthermore, sparsity constraints encourage auxiliary non-physical terms to maintain negligible coefficients compared to dominant, mechanism-aligned terms. This selective stability demonstrates ability of COSINE to identify parsimonious, meaningful terms.
% Overall, COSINE balances accuracy and interpretability by stabilizing core primitives while suppressing spurious nonlinearities.

\subsection{Results on Real-World Epidemic Data}

We apply COSINE to COVID-19 data from four U.S. states (\cref{appd:emp_exp}) to jointly infer latent transmission structures and governing equations. The inferred graphs (\cref{fig:epi_results}a) reveal pronounced regional heterogeneity: Arizona and Connecticut exhibit dense connectivity driven by high mobility and metropolitan proximity, while Illinois and Michigan are significantly sparser, reflecting urban-rural hierarchies and localized clustering. Dynamically, the Message modules across all regions universally identify multiplicative terms ($x_i \cdot x_j$), which align with the classical mass-action effect of contact-based transmission. This suggests that cross-regional spreading intensity is consistently determined by the coupling between local and neighboring infection levels.

The Update modules further distinguish distinct regional transmission regimes (\cref{fig:epi_results}b). In Arizona and Connecticut, the discovered terms frequently involve external pressure expressions such as $x \cdot h$, indicating that local dynamics are highly vulnerable to the accumulation of incoming infections. Conversely, the dynamics in Illinois and Michigan are dominated by local nonlinearities---such as logistic growth $x(1-x)$ and negative feedback terms---capturing endogenous spreading patterns and the impact of localized intervention measures. These findings demonstrate COSINE's capacity to extract mechanistic, physics-informed insights that can support differentiated public health decision-making in complex real-world systems.

\subsection{Ablation Study}

\begin{figure}[!t]
\vskip 0.1in
\begin{center}
\includegraphics[width=0.9\linewidth]{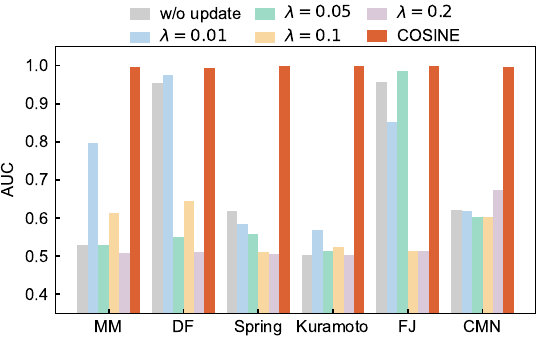}
\caption{Abalation of LLM-guided evolution. COSINE vs. fixed library (w/o update) and threshold-based pruning ($\lambda$).}
\label{fig:llm-update}
\end{center}
\vskip -0.1in
\end{figure}

% \paragraph{Effectiveness of LLM-Guided Evolution.}
% As shown in \cref{fig:llm-update}, we evaluate the necessity of the LLM update mechanism by comparing COSINE against a baseline using a fixed initial library (\emph{w/o LLM update}). COSINE consistently achieves superior structural inference AUC across all six systems, with the performance gap being particularly pronounced in complex nonlinear regimes such as \emph{Kuramoto}, \emph{MM}, and \emph{CMN}. This highlights that one-shot symbolic generation is often insufficient for capturing unknown mechanisms; in contrast, COSINE's closed-loop ``generation-evaluation-update'' process allows the LLM to act as a dynamic symbolic editor, surmounting the expression bottlenecks of static libraries by aligning symbolic hypotheses with underlying physical laws via training feedback.

% \begin{figure}[!ht]
% \vskip 0.1in
% \begin{center}
% \includegraphics[width=\linewidth]{figs/llm_update_er_only.pdf}
% \caption{Ablation analysis of structural inference AUC on ER graphs, comparing COSINE with a fixed initial library version.}
% \label{fig:llm-update}
% \end{center}
% \vskip -0.1in
% \end{figure}

\paragraph{Ablation of LLM-Guided Evolution.}
% As illustrated in \cref{fig:llm-update}, we evaluate the necessity of the LLM-guided evolution mechanism by comparing COSINE against static libraries (\emph{w/o update}) and threshold-based pruning strategies ($\lambda$). While COSINE consistently achieves near-perfect performance (AUC $\approx 1.0$) across all systems, baselines with fixed or passively filtered libraries exhibit significant performance drops and inconsistency, particularly in complex regimes like \emph{MM} and \emph{CMN}. This disparity underscores the ``closed-world'' bottleneck of static modeling, where structural inference is fundamentally capped by the initial quality of the basis set. Unlike numerical heuristics that provide only passive weight-based filtering, the LLM-guided module treats the library as a malleable hypothesis space, performing mechanistic reasoning to actively prune redundant terms and augment promising nonlinearities based on inner-loop feedback. 
As illustrated in \cref{fig:llm-update}, we evaluate the necessity of the LLM-guided evolution mechanism by comparing COSINE against static libraries (\emph{w/o update}) and threshold-based pruning strategies ($\lambda$). While COSINE consistently achieves near-perfect performance (AUC $\approx 1.0$) across all systems, baselines with fixed or passively filtered libraries exhibit significant performance drops and inconsistency, particularly in complex regimes like \emph{MM} and \emph{CMN}. This disparity underscores the ``closed-world'' bottleneck of static modeling, where structural inference is fundamentally capped by the initial quality of the basis set. Unlike numerical heuristics that provide only passive weight-based filtering, in sharp contrast, the LLM-guided module treats the library as a malleable hypothesis space, performing mechanistic reasoning to actively prune redundant terms and augment promising nonlinearities based on inner-loop feedback.

\begin{figure}[!t]
    \vskip 0.1in
    \begin{center}
    \centerline{\includegraphics[width=0.9\linewidth]{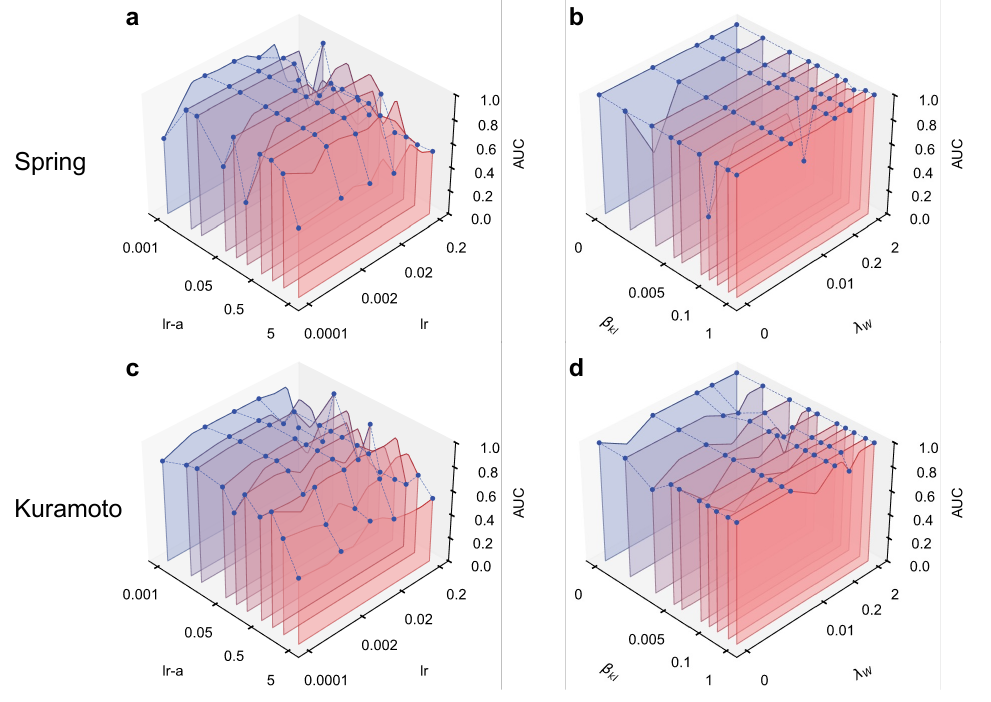}}
    % \caption{Hyperparameter sensitivity analysis of COSINE. Panels (a) and (b) show the results for the Spring system, while (c) and (d) show those for the Kuramoto system. Panels (a) and (c) illustrate the sensitivity to learning rates ($\text{lr}_a$ vs. $\text{lr}$), and panels (b) and (d) depict the impact of regularization weights ($\beta_{\mathrm{KL}}$ vs. $\lambda_W$). Performance is measured by structural inference AUC.}
    \caption{Hyperparameter sensitivity analysis of COSINE for Spring (a--b) and Kuramoto (c--d). (a, c) $\text{lr}_a$ vs. $\text{lr}$; (b, d) $\beta_{\mathrm{KL}}$ vs. $\lambda_W$. Performance: relational inference AUC.}
    \label{fig:sensitivity}
    \end{center}
    \vskip -0.1in
\end{figure}

\paragraph{Impact of LLM Scale on Performance.}
% We evaluate how the scale of the LLM used in the outer loop affects the downstream structural recovery performance. \cref{tab:llm_scale_performance} reports the AUC across representative dynamical systems.

\begin{table}[!t]
\caption{Performance (AUC) of COSINE with different LLMs. Models are ordered by increasing parameter scale.}
\label{tab:llm_scale_performance}
\vskip 0.1in
\begin{center}
\begin{small}
\begin{sc}
\renewcommand{\arraystretch}{1.0}
\setlength{\tabcolsep}{6pt}
\begin{tabular}{l c c c c c}
\toprule
Model & Params & Spr & Kura & MM & CMN \\
\midrule
DeepSeek-R1 & 8B   & 1.00 & 1.00 & 0.53 & 0.51 \\
Qwen3       & 8B   & 1.00 & 1.00 & 0.54 & 0.66 \\
\midrule
DeepSeek-R1 & 14B  & 1.00 & 0.99 & 0.88 & 1.00 \\
GPT-OSS     & 20B  & 1.00 & 0.99 & 1.00 & 1.00 \\
\midrule
DeepSeek-R1 & 32B  & 1.00 & 1.00 & 0.94 & 1.00 \\
Qwen3       & 32B  & 1.00 & 0.83 & 0.99 & 0.90 \\
\midrule
Qwen2.5     & 72B  & 0.99 & 1.00 & 1.00 & 1.00 \\
GPT-OSS     & 120B & 1.00 & 1.00 & 0.96 & 1.00 \\
\bottomrule
\end{tabular}
\end{sc}
\end{small}
\end{center}
\vskip -0.1in
\end{table}

As shown in \cref{tab:llm_scale_performance}, COSINE is only weakly sensitive to LLM scale on simpler dynamics: Spring stays near ceiling across all models, and Kuramoto remains close to $1.00$ once sinusoidal primitives are available. The scale effect becomes clear for higher-complexity nonlinear systems (MM and CMN), where 8B models underperform and mid-to-large models (e.g., 14B+ and Qwen2.5-72B) stabilize at strong AUC. Overall, LLM size matters primarily when the dynamics require richer compositional primitives; otherwise, COSINE remains robust with smaller models.

\paragraph{Hyperparameter Sensitivity Analysis.} 
We evaluate robustness of COSINE across various hyperparameters including learning rates of graph and weights (Group 1: $\text{lr}_a, \text{lr}$) and regularization weights of graph and basis (Group 2: $\beta_{\mathrm{KL}}, \lambda_W$). As shown in \cref{fig:sensitivity}, structural inference exhibits a distinct ``performance plateau'' and a well-defined ``failure boundary'' regarding learning rates; COSINE consistently achieves near-perfect AUC ($\approx 1.0$) within optimal ranges, though excessively high rates trigger sharp declines due to optimization instability. In contrast, the model demonstrates remarkable robustness to regularization parameters, maintaining stable high AUC even under substantial sparsity penalties ($\lambda_W$ up to $2.0$). This stability suggests that the framework effectively prunes redundant terms without compromising structural recovery, highlighting the synergy between LLM-guided symbolic selection and inner-loop differentiable optimization.

\subsection{Efficiency Analysis}

\begin{figure}[!t]
\vskip 0.1in
\begin{center}
\centerline{\includegraphics[width=\linewidth]{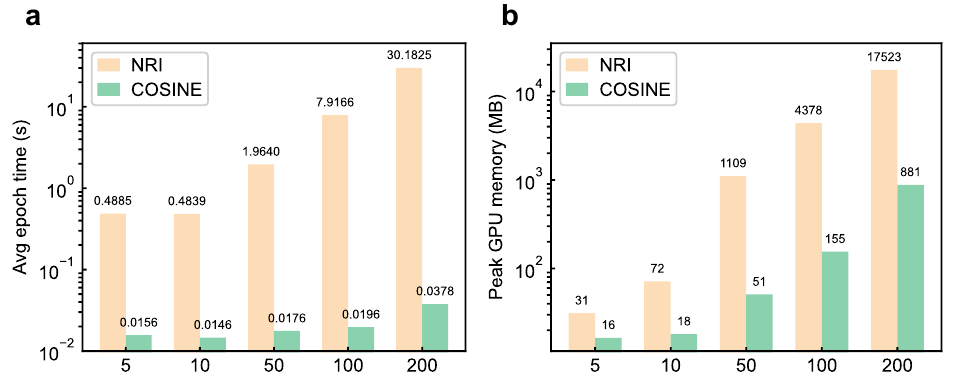}}
\caption{Computational efficiency and memory scaling comparison between COSINE and NRI (batch size 64, hidden dimension 256) across different network sizes $N \in \{5, 10, 50, 100, 200\}$. (a) Average epoch time, (b) Peak GPU memory usage. Note the logarithmic scale on the vertical axes.}
\label{fig:efficiency}
\end{center}
\vskip -0.1in
\end{figure}

% \paragraph{Computational Efficiency and Scalability.} 
As shown in \cref{fig:efficiency}, COSINE consistently outperforms NRI in training time and GPU memory across all tested scales. This efficiency stems from replacing heavy neural networks with a lightweight sparse linear regression framework over a symbolic basis library, which bypasses the extensive parameter optimization required for high-dimensional neural representations. Consequently, COSINE demonstrates superior scalability up to $N=200$ nodes, maintaining a parsimonious memory footprint that scales gracefully compared to the steep overhead of conventional neural models. These results underscore that decoupling symbolic search from differentiable optimization provides a highly efficient pathway for structure and mechanism discovery in large-scale networked dynamical systems.

\section{Limitations}
Despite strong performance, COSINE has limitations. Complex dynamics can place stronger demands on the outer-loop LLM. Our sparse linear regression over a learned library improves tractability, but the expressivity of sparse symbolic regression is bounded and may require richer compositional primitives. Under finite noisy data, symbolic equation discovery may be non-identifiable, with near-equivalent terms blurring mechanistic conclusions.

\section{Conclusion}

In this paper, we presented COSINE, a differentiable framework for interpretable relational inference that jointly learns latent interaction graphs and sparse dynamical equations. COSINE models dynamics as sparse combinations over message/update basis libraries and employs an LLM to iteratively refine these libraries based on optimization feedback. Beyond physics-inspired systems, the message--update factorization provides a general template for \emph{networked dynamics} across domains, where interactions may encode biochemical regulation, social influence, mobility-driven transmission, or other mechanisms. Experiments on synthetic and real-world systems show that COSINE achieves strong structural recovery while yielding mechanism-aligned terms and favorable efficiency and scalability.

% one-step training and pairwise message--update factorization may miss long-horizon or higher-order mechanisms,
\section*{Impact Statement}

% Authors are \textbf{required} to include a statement of the potential broader
% impact of their work, including its ethical aspects and future societal
% consequences. This statement should be in an unnumbered section at the end of
% the paper (co-located with Acknowledgements -- the two may appear in either
% order, but both must be before References), and does not count toward the paper
% page limit. In many cases, where the ethical impacts and expected societal
% implications are those that are well established when advancing the field of
% Machine Learning, substantial discussion is not required, and a simple
% statement such as the following will suffice:

This paper presents work whose goal is to advance the field of Machine
Learning. There are many potential societal consequences of our work, none of 
which we feel must be specifically highlighted here.

% The above statement can be used verbatim in such cases, but we encourage
% authors to think about whether there is content which does warrant further
% discussion, as this statement will be apparent if the paper is later flagged
% for ethics review.


\bibliography{example_paper}
\bibliographystyle{icml2026}

%%%%%%%%%%%%%%%%%%%%%%%%%%%%%%%%%%%%%%%%%%%%%%%%%%%%%%%%%%%%%%%%%%%%%%%%%%%%%%%
%%%%%%%%%%%%%%%%%%%%%%%%%%%%%%%%%%%%%%%%%%%%%%%%%%%%%%%%%%%%%%%%%%%%%%%%%%%%%%%
% APPENDIX
%%%%%%%%%%%%%%%%%%%%%%%%%%%%%%%%%%%%%%%%%%%%%%%%%%%%%%%%%%%%%%%%%%%%%%%%%%%%%%%
%%%%%%%%%%%%%%%%%%%%%%%%%%%%%%%%%%%%%%%%%%%%%%%%%%%%%%%%%%%%%%%%%%%%%%%%%%%%%%%
\newpage
\appendix
\onecolumn

\section{Details of Method}

\subsection{Training Process of COSINE}\label{appd:train_process}
The overall training procedure of COSINE is summarized in \cref{alg:COSINE-training}. The framework follows a nested co-optimization structure consisting of an outer symbolic discovery loop and an inner joint optimization loop.

\paragraph{Inner Loop: Joint Structure--Equation Optimization.} Within each outer round, with the basis function library $\Theta$ fixed, the model seeks to simultaneously recover the network structure and the governing dynamical equations. The graph generator parameters $\bm{\Psi}$ and the sparse regression coefficients $\mathbf{W}$ are jointly updated via gradient descent to minimize the total loss $\mathcal{L}_{\text{total}}$. This inner loop leverages the differentiability of the Gumbel--Softmax edge sampling and the linear basis expansion to efficiently navigate the high-dimensional parameter space, identifying the most probable interaction structure and the corresponding sparse weights that characterize the system's evolution.

\paragraph{Outer Loop: LLM-Guided Library Evolution.} After the inner loop converges, the LLM $\mathcal{M}$ acts as a symbolic supervisor to refine the library $\Theta$. It analyzes the training results---including validation performance scores, basis term weights (importance measures), and residual patterns---to decide which terms should be pruned due to negligible contributions or augmented to capture remaining nonlinearities. This iterative process of hypothesis generation, evaluation, and reflection allows COSINE to escape the "closed-world" limitation of traditional sparse regression, progressively evolving towards a parsimonious yet accurate symbolic representation of the underlying mechanisms.

\begin{algorithm*}[h]
    \caption{Overall Training Procedure of COSINE}
    \label{alg:COSINE-training}
\begin{algorithmic}
    \STATE {\bfseries Input:} Node trajectories $\mathbf{X}$, initial library $\Theta^{(0)}$, max rounds $R$, max epochs $E$, weights $\beta_{\mathrm{KL}}, \lambda_{W}$
    \STATE {\bfseries Output:} Learned adjacency $\mathbf{A}$, coefficients $\mathbf{W}$, discovered library $\Theta^{\star}$
    \STATE Initialize structure parameters $\bm{\Psi}$ and regression weights $\mathbf{W}$
    \STATE Set best library $\Theta^{\star} \leftarrow \Theta^{(0)}$ and best validation loss $\mathcal{L}_{\mathrm{val}}^{\star} \leftarrow +\infty$
    \FOR{$r = 0$ {\bfseries to} $R-1$}
        \STATE Fix current library $\Theta^{(r)}$
        \FOR{$e = 1$ {\bfseries to} $E$}
            \STATE Sample mini-batch $(\mathbf{x}^{t}, \mathbf{x}^{t+1})$ from $\mathbf{X}$
            \STATE Compute soft adjacency $\mathbf{A}^{\mathrm{soft}}$ from $\bm{\Psi}$ using Gumbel--Softmax
            \STATE Compute message and update features using $\Theta^{(r)}$ and predict $\hat{\mathbf{x}}^{t+1}$
            \STATE Compute losses $\mathcal{L}_{\mathrm{nll}}, \mathcal{L}_{A}, \mathcal{L}_{W}$ and total loss $\mathcal{L}_{\mathrm{total}}$
            \STATE Update $\bm{\Psi}$ and $\mathbf{W}$ by gradient descent on $\mathcal{L}_{\mathrm{total}}$
        \ENDFOR
        \STATE Evaluate validation loss $\mathcal{L}_{\mathrm{val}}^{(r)}$ and record mean absolute weights for each basis term
        \IF{$\mathcal{L}_{\mathrm{val}}^{(r)} < \mathcal{L}_{\mathrm{val}}^{\star}$}
            \STATE $\mathcal{L}_{\mathrm{val}}^{\star} \leftarrow \mathcal{L}_{\mathrm{val}}^{(r)}$, $\Theta^{\star} \leftarrow \Theta^{(r)}$, and store parameters $\bm{\Psi}, \mathbf{W}$
        \ENDIF
        \STATE Build LLM prompt from $\Theta^{\star}$, importance statistics, and residual patterns
        \STATE Query LLM $\mathcal{M}$ to obtain a modified library $\widetilde{\Theta}$ via pruning and augmentation
        \STATE Set $\Theta^{(r+1)} \leftarrow \widetilde{\Theta}$ and reinitialize $\mathbf{W}$; optionally warm-start $\bm{\Psi}$
    \ENDFOR
    \STATE {\bfseries return} $\Theta^{\star}, \bm{\Psi}, \mathbf{W}$
\end{algorithmic}
\end{algorithm*}

\subsection{Basis Search Implementation Details}\label{appd:basis_search}

The outer-loop optimization of COSINE can be viewed as a search process over the discrete space of symbolic structures via \emph{basis library editing}. Each round of LLM-guided refinement, involving the augmentation or pruning of basis terms, essentially alters the feasible subspace of dynamical expressions. While sophisticated search algorithms such as Monte Carlo Tree Search (MCTS) could be employed to explicitly manage a multi-branch search tree and balance exploration versus exploitation, such approaches entail significant computational overhead due to recursive branching and the requirement for multiple full model training and evaluation cycles. In settings that require frequent LLM interactions, MCTS introduces unnecessary complexity that may not be justified by the marginal gains.

Instead, we adopt a more lightweight and reproducible \textbf{best-so-far} strategy. This approach maintains a single current optimal basis library and only accepts a proposed candidate if it yields a strictly \textbf{lower} validation loss. Formally, let $\Theta^{(r)}$ denote the basis library at outer iteration $r$. Its performance is evaluated by the inner-loop joint optimization (of structural parameters $\bm{\Psi}$ and regression coefficients $\mathbf{W}$) as the total validation loss:
\begin{equation}
\mathcal{L}_{\mathrm{val}}(\Theta)=\mathcal{L}_{\mathrm{total}}\quad \text{(evaluated on the validation set)}.
\end{equation}

We maintain the optimal library $\Theta^{\star}$ identified throughout the search:
\begin{equation}
\Theta^{\star}=\arg\min_{\Theta\in\{\Theta^{(0)},\dots,\Theta^{(r)}\}}\mathcal{L}_{\mathrm{val}}(\Theta).
\end{equation}
In each subsequent outer iteration, $\Theta^{\star}$—along with its corresponding importance statistics and training residual patterns—is provided as core context in the prompt to generate a candidate library $\widetilde{\Theta}$. Following a round of inner-loop joint training, we obtain $\mathcal{L}_{\mathrm{val}}(\widetilde{\Theta})$. If $\mathcal{L}_{\mathrm{val}}(\widetilde{\Theta}) < \mathcal{L}_{\mathrm{val}}(\Theta^{\star})$, we update $\Theta^{\star} \leftarrow \widetilde{\Theta}$. Otherwise, the candidate is rejected, and $\Theta^{\star}$ is preserved for the next iteration.

This strategy serves as a "greedy update with rejection" mechanism rather than a full tree search. By not explicitly maintaining multi-branch states, we mitigate the risk of performance degradation arising from LLM pruning errors or the introduction of noisy augmentation terms. The primary advantages of this approach are its controllable computational cost and implementation simplicity, which effectively leverages the strong feedback signals from numerical optimization and sparsity constraints to steer the symbolic search towards physically consistent and parsimonious mechanism regions.

\subsection{LLM Details}\label{appd:llm_details}

For the core experiments presented in this work, we utilize the \textbf{GPT-OSS (20B)} model as the default Large Language Model. This model is deployed locally using the Ollama framework and runs on an NVIDIA RTX 4090 GPU, ensuring data privacy and low-latency feedback during the outer-loop library evolution. For the analysis of real-world epidemic data, we employ the \textbf{MiMo-V2-Flash} model provided via the OpenRouter API.

% It should be noted that while these two models serve as the primary backends, our framework is model-agnostic. In specific ablation studies (such as the parameter scale comparison reported in Table~\ref{tab:llm_scale_performance_single}), we further evaluate various LLMs across different parameter scales to investigate the relationship between model capacity and symbolic discovery performance.

\subsection{Prompt Details}

In the COSINE framework, we utilize three distinct prompt templates to coordinate the interaction between numerical optimization and LLM-based symbolic discovery: the \emph{System Prompt}, the \emph{Initial Basis Prompt}, and the \emph{Refine Basis Prompt}. The \emph{System Prompt} establishes the foundational operational framework, including strict variable scoping, tensor dimension constraints, and numerical stability expectations. The \emph{Initial Basis Prompt} is designed for the zero-shot generation of the starting candidate library. The \emph{Refine Basis Prompt} incorporates multi-faceted feedback from the inner-loop training—including validation performance, coefficient importance statistics, and residual patterns—to guide the iterative evolution of the library.

\subsubsection{System Prompt}
To ensure the symbolic consistency and numerical stability of the generated basis functions, we provide the LLM supervisor with a comprehensive system prompt. This prompt establishes a strict operational framework, defining available tensor variables, mathematical primitives, and dimension-matching constraints for both message-passing and node-update modules. By explicitly outlining forbidden operations and stability expectations (e.g., handling singularities via small epsilon values), the prompt guides the LLM to propose parsimonious yet physically grounded candidate terms while adhering to the differentiable PyTorch-based execution environment.

The complete system prompt used to initialize and guide the basis library evolution is presented in \cref{lst:system_prompt}.

\lstset{
  basicstyle=\ttfamily\scriptsize,
  breaklines=true,
  frame=single,
  captionpos=b,
  commentstyle=\color{gray},
  keywordstyle=\bfseries,
  showstringspaces=false
}

\begin{lstlisting}[caption={System Prompt for the LLM Basis Library Supervisor}, label={lst:system_prompt}]
You are an expert in discovering dynamical systems and their interaction structures using sparse symbolic regression.

Your Task:
- Act as an expert in dynamical systems, sparse regression, and PyTorch expressions.
- Only design / refine the basis function library.

Output Requirements (MUST be strictly followed):
1) Respond with a single JSON object and nothing else (no markdown or explanatory text).
2) The JSON must contain exactly two array fields: "message_terms" and "update_terms".
   - Functions in message_terms are used to model interactions between nodes.
   - Functions in update_terms are used to model each node's own state update.
3) Each item must contain exactly three fields: {"name", "expr", "type"}, where type in {"vector", "scalar"}.
   - The field name must be exactly "expr".
   - No extra fields are allowed, no empty field is allowed.
4) Within the same stream, names must be unique.

Available tensors and helpers:
- Message stream only:
  xi, xj shape: [B, N, N, D]; diff = xj - xi
- Update stream only:
  x in [B, N, D]; h in [B, N, D]; deg in [B, N, 1]
- Shared helpers: torch, F, pi, inf

Strict variable scoping rules (NO EXCEPTIONS):
- Expressions in message_terms may only use: xi, xj, diff (and shared helpers); never use x / h / deg.
- Expressions in update_terms may only use: x, h, deg (and shared helpers); never use xi / xj / diff.
- Never introduce new symbols.

Shape and type rules:
- Outputs of message_terms must be broadcastable to:
  - Vector terms: [B, N, N, D]
  - Scalar terms: [B, N, N, 1]
- Outputs of update_terms must be:
  - Vector terms: [B, N, D]
  - Scalar terms: [B, N, 1]
- Scalar terms MUST keep a trailing singleton dimension (..., 1).

Strictly forbidden patterns:
- In update_terms: NEVER reference variables such as "msg" or "message"; only "h" exists.
- In update_terms: NEVER create an extra N dimension (no [B, N, N, *]).
- NEVER use reshape/reordering operations:
  view / reshape / permute / transpose / unsqueeze / squeeze.
- Forbidden constructors / stacking operations:
  torch.tensor / torch.zeros / torch.ones / torch.cat / torch.stack.
  If constants are needed, use torch.zeros_like(x) or torch.ones_like(x).

Expression rules:
- PyTorch-only expressions (torch.* / F.* and basic arithmetic).
- No numpy, no math.*, no control flow.
- No attributes containing "__" or starting with an underscore.

Recommended primitives and templates (allowed if they obey scoping + shape rules):
- Polynomial / power: x, x*x, torch.pow(x, 2), torch.pow(x, 3), torch.sqrt(torch.clamp(x, min=0)).
- Trigonometric: torch.sin(x), torch.cos(x), torch.tan(x).
- Exponential / log: torch.exp(x), torch.log(torch.clamp(x, min=1e-6)).
- Activations / saturation: torch.sigmoid(x), torch.tanh(x), F.relu(x), F.softplus(x).
- Rational / fractional (numerically stable):
  - Inverse: 1.0 / (x + 1e-6)
  - Saturation (Hill-like): x / (1.0 + torch.abs(x))
  - Squared saturation: x / (1.0 + x*x)
  Use small eps (e.g., 1e-6) to avoid division-by-zero.

Numerical stability expectations:
- Avoid singularities: add eps in denominators; clamp inputs to log/sqrt.
- Prefer bounded transforms (tanh/sigmoid) when exploring complex nonlinearity.
- Occam's Razor: Prefer simpler expressions. Only introduce complexity (e.g., rational functions, high-order polynomials) if simple linear or low-order terms fail to capture the dynamics.
- expr must be a single valid Python expression parseable by:
  ast.parse(expr, mode="eval").
- No placeholders are allowed inside expr.
- Expression length must be less than 500 characters.
- Numerical safety: when dividing by a potentially small quantity, add + 1e-6.
- torch.norm MUST be written as:
  torch.norm(x, dim=-1, keepdim=True)
  and NEVER as torch.norm(x, -1, ...).

Multi-channel (D > 1) channel slicing rules (CRITICAL):
- NEVER use x[..., c] / xj[..., c] (this drops the last dimension).
- ALWAYS use x[..., c:c+1] / xj[..., c:c+1], and never write c:c.
- Cross-channel interactions should be designed as scalar terms (then broadcast to D).

Budget constraints:
- Each call specifies max_terms.
- Must satisfy:
  - len(message_terms) <= max_terms
  - len(update_terms) <= max_terms
  - message_terms and update_terms must each contain at least one term.

Minimal example (FORMAT ONLY):
{
  "message_terms": [
    {"name": "diff", "expr": "xj - xi", "type": "vector"}
  ],
  "update_terms": [
    {"name": "h", "expr": "h", "type": "vector"}
  ]
}
\end{lstlisting}

\subsubsection{Init Basis Prompt}
The \emph{Initial Basis Prompt} (shown in \cref{lst:init_prompt}) is used to initialize the symbolic library $\Theta^{(0)}$. It instructs the LLM to prioritize parsimony and core physical primitives (e.g., linear differences and basic state terms) before exploring higher-order nonlinearities. The prompt incorporates several key environment variables:
\begin{itemize}
    \item \texttt{D}: Represents the feature dimension of node states.
    \item \texttt{system description}: Provides a natural language functional description of the underlying system. Notably, this variable is \textbf{exclusively} utilized for the real-world epidemic data experiments to provide domain-specific grounding; in all synthetic benchmarks, it is omitted.
    \item \texttt{max\_terms}: Specifies the maximum number of basis functions permitted for each of the message and update streams, ensuring complexity control.
\end{itemize}

\begin{lstlisting}[caption={Initial Basis Prompt Template}, label={lst:init_prompt}]
You are designing the initial basis function library for COSINE dynamics discovery.

Follow all hard rules in `system_prompt.txt`. This file only adds initialization-phase guidance.

Problem snapshot:
- State dimension D = {feature_dim}
- System description: {description}
- Budget: at most {max_terms} terms in each stream

Initialization strategy (Simplicity First):
1) Message function (message_terms):
    - Start with simple and plausible interaction mechanisms (e.g., linear difference `xj - xi`).
2) Update function (update_terms):
    - Include core baseline terms (e.g., `x`, `h`) to represent basic self-dynamics and neighbor influence.
    - Keep terms low-order (linear or quadratic) initially.
3) Multi-channel (D>1):
    - Use simple channel-wise terms before exploring cross-channel interactions.

Advanced Exploration (if the system description suggests high complexity):
- If the system is known to be highly non-linear (e.g., biological, chemical), you may include terms from the following categories:
  - Trigonometric: sin(.) / cos(.) on diff or state.
  - Activation/saturation: tanh(.) / sigmoid(.).
  - Rational: x/(1+|x|) or 1/(x+eps).

Graph-specific normalization (update_terms only):
- If the system likely depends on node degree, include `h / (deg + 1e-6)`.

Avoid near-duplicate terms; keep names stable and unique.
\end{lstlisting}

\subsubsection{Refine Basis Prompt}
The \emph{Refine Basis Prompt} (shown in \cref{lst:refine_prompt}) governs the iterative evolution of the symbolic library during the outer-loop optimization. In alignment with the "best-so-far" strategy described in Appendix B.2, this prompt acts as a symbolic editor that modifies the current optimal library based on numerical feedback. It incorporates two types of historical context:
\begin{itemize}
    \item \texttt{recent\_rounds}: A summary of the symbolic libraries and training results (NLL and coefficient importance statistics) from the most recent 3 iterations. This allows the LLM to identify persistent low-weight terms for pruning or recognize when the search has stalled in a specific functional subspace.
    \item \texttt{best\_round\_details}: The complete specification of the global optimal library $\Theta^{\star}$ identified thus far. This includes naming, mathematical expressions, and the learned importance of each term, serving as the definitive baseline for subsequent mutations and complexity escalation.
\end{itemize}

\begin{lstlisting}[caption={Refine Basis Prompt Template}, label={lst:refine_prompt}]
You are optimizing basis function selection based on training feedback. Reasoning: High.

All hard constraints (JSON format, tensors, scoping, shapes, forbidden operations, etc.) are specified in system_prompt.txt and must be strictly followed.

Inputs provided with each call include:
- Recent 3 rounds (each round has nll + Message terms + Update terms; each term includes name, mean|w|, expr):
{recent_rounds}
- Best round so far (nll + Message terms + Update terms; each term includes name, mean|w|, expr):
{best_round_details}

=== DECISION LOGIC ===

Step 1: Optimize and explore based on historical feedback, using the best round's basis functions as the starting point.

Step 2: Weight-guided refinement and Complexity Escalation
Use your reasoning to refine the library based on weights and NLL:
- Retain high-weight terms: If a simple term (e.g., linear) has a high weight, KEEP it. Do not replace it with a complex version unless the NLL is high and progress has stalled.
- Prune low-weight terms: Remove terms that consistently show near-zero weights.
- Progressive Complexity: introduce more complex terms (e.g., rational, activations, high-order polynomials) if the current simple terms are insufficient to reduce the NLL.

Step 2b: Structure-guided mutation (Escalate only when needed)
If the fit plateaus or NLL remains high, try MUTATING weak terms into slightly more complex forms:
- Escalation Path: Linear -> Polynomial (x^2) -> Non-linear (sin/tanh) -> Rational (x/(1+x)).
- Degree normalization (update_terms only): If neighbor influence `h` is weak, try `h/(deg+1e-6)`.
- Interaction nonlinearity (message_terms): If `diff` is weak, try `sin(diff)` or `tanh(diff)`.

=== KEY INSIGHT: Occam's Razor ===
Prefer the simplest expression that explains the data. A simple term with a high weight is much better than a complex term with a similar weight.

=== BUDGET ===
- Budget: at least 1 term per stream, at most {max_terms} terms per stream.
- Ensure the basis function library can adequately and clearly express the dynamical equations.
- Critical requirement: within the same stream, every term name must be UNIQUE.
\end{lstlisting}

\section{Details of Synthetic Experiment}\label{appd:syn_exp}

\subsection{Dynamical Systems}
Details of the six representative classes of networked dynamical systems used in experiments:

\subsubsection{Michaelis--Menten kinetics (MM)}
\begin{figure}[ht]
    \vskip 0.2in
    \begin{center}
    \includegraphics[width=0.8\linewidth]{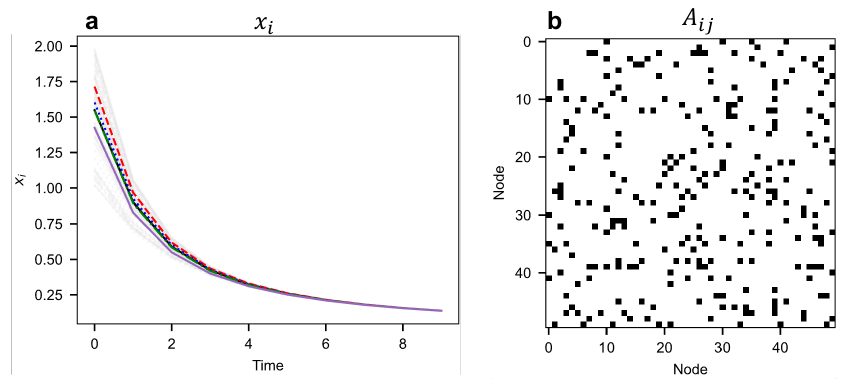}
    \caption{Visual illustration of Michaelis--Menten kinetics. \textbf{a}, Node activities (with the trajectories of the first 5 nodes highlighted by colored lines). \textbf{b}, Adjacent matrix of the network.}
    \label{fig:app_mm}
    \end{center}
    \vskip -0.2in
\end{figure}

Michaelis--Menten (MM) kinetics~\citep{karlebach2008modelling} is a fundamental continuous-time nonlinear model widely used to describe gene-regulatory interactions and biochemical reaction networks. In this model, the state evolution of each node is governed by a combination of basal production, nonlinear cooperative interactions, and first-order degradation. The dynamic equation is given by:
\begin{equation}
    \dot x_{i} = -x_{i} + \sum_{j\in\mathcal{N}(i)}\frac{A_{ij}}{k_{i}}\,\frac{x_{j}}{1+x_{j}}
\end{equation}
where $x_i$ represents the activity level of node $i$, and $k_i = \sum_j A_{ij}$ is the in-degree of node $i$. This formulation assumes unit degradation rate and saturation constants, with normalized interaction strengths across neighbors. This model effectively captures the saturation effects inherent in biological signaling. \cref{fig:app_mm} illustrates the representative trajectories and the underlying ground-truth adjacency matrix for the Michaelis--Menten system.

\subsubsection{Diffusion (Diff)}
\begin{figure}[ht]
    \vskip 0.2in
    \begin{center}
    \includegraphics[width=0.8\linewidth]{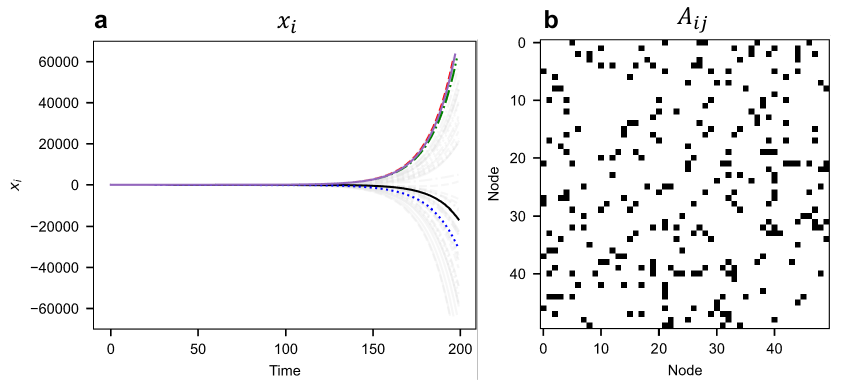}
    \caption{Visual illustration of Diffusion dynamics. \textbf{a}, Node activities (with the trajectories of the first 5 nodes highlighted by colored lines). \textbf{b}, Adjacent matrix of the network.}
    \label{fig:app_diff}
    \end{center}
    \vskip -0.2in
\end{figure}

The diffusion process represents a classic continuous-time linear mechanism for conservative flow or heat distribution over a network. It assumes that the rate of change of a node's state is proportional to the difference between its state and those of its neighbors. The global system dynamics are expressed as:
\begin{equation}
    \dot x = -\beta L x,\quad L=D_{\text{in}}-A
\end{equation}
where $\beta > 0$ denotes the global diffusion rate (or thermal conductivity). In our experiments, we set $\beta = 1.0$. $L$ is the directed graph Laplacian defined as $L = \mathbf{D}_{\text{in}} - \mathbf{A}$, where $\mathbf{D}_{\text{in}}$ is the in-degree diagonal matrix and $\mathbf{A}$ is the adjacency matrix. This model is a cornerstone for studying spreading processes and consensus algorithms. \cref{fig:app_diff} visualizes the diffusing states alongside the network.

\subsubsection{Network of Springs (Spring)}
\begin{figure}[ht]
    \vskip 0.2in
    \begin{center}
    \includegraphics[width=0.8\linewidth]{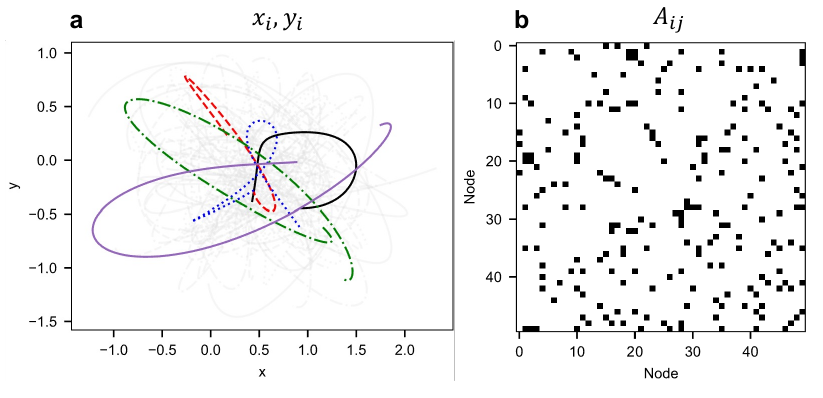}
    \caption{Visual illustration of Spring dynamics. \textbf{a}, Node activities (with the trajectories of the first 5 nodes highlighted by colored lines). \textbf{b}, Adjacent matrix of the network.}
    \label{fig:app_spring}
    \end{center}
    \vskip -0.2in
\end{figure}

The network of springs (Spring or Spr) is a continuous-time second-order dynamical system derived from classical mechanics. It treats each node as a unit mass coupled to its neighbors by linear springs satisfying Hooke’s law, often including a global damping effect. The dynamics follow:
\begin{equation}
    \ddot{\mathbf{p}}_{i} = -\gamma\dot{\mathbf{p}}_{i} - k\sum_{j\in\mathcal{N}(i)}A_{ij}(\mathbf{p}_{i}-\mathbf{p}_{j})
\end{equation}
where $\mathbf{p}_i$ denotes the position of node $i$, $k$ is the spring constant, and $\gamma$ is a global damping coefficient. \cref{fig:app_spring} illustrates the representative trajectories and the underlying ground-truth adjacency matrix for the Spring system.

\subsubsection{Kuramoto Model (Kuramoto)}
\begin{figure}[ht]
    \vskip 0.2in
    \begin{center}
    \includegraphics[width=0.8\linewidth]{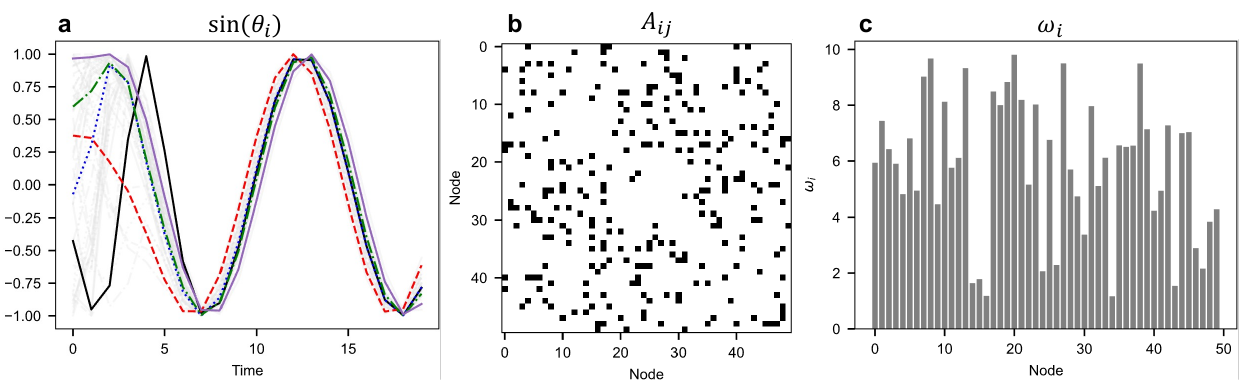}
    \caption{Visual illustration of Kuramoto model. \textbf{a}, Node activities (with the trajectories of the first 5 nodes highlighted by colored lines). \textbf{b}, Adjacent matrix of the network. \textbf{c}, Natural frequencies $\omega_i$ of each node.}
    \label{fig:app_kura}
    \end{center}
    \vskip -0.2in
\end{figure}

The Kuramoto model~\citep{kuramoto1975self} is a canonical framework for studying synchronization phenomena in populations of coupled oscillators. It describes the evolution of the phases of $N$ oscillators that interact through a nonlinear sine-coupling term. The governing equation is:
\begin{equation}
    \dot\phi_{i} = \omega_{i} + \sum_{j\in\mathcal{N}(i)}A_{ij}\kappa\sin(\phi_{j}-\phi_{i})
\end{equation}
where $\phi_i$ is the phase of oscillator $i$, $\omega_i$ is its natural frequency, and $\kappa$ denotes the coupling strength. \cref{fig:app_kura} visualizes the phase evolution, network, and natural frequencies.

\subsubsection{Friedkin--Johnsen Dynamics (FJ)}
\begin{figure}[ht]
    \vskip 0.2in
    \begin{center}
    \includegraphics[width=0.8\linewidth]{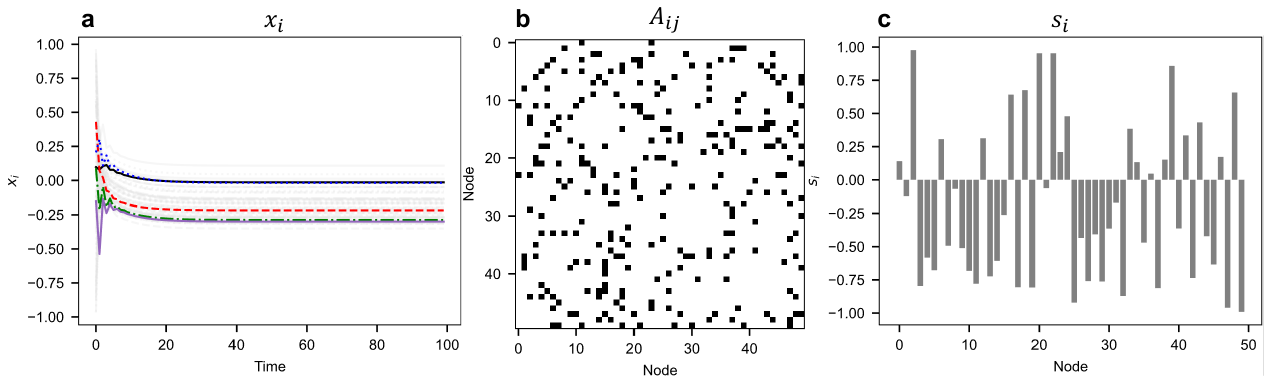}
    \caption{Visual illustration of Friedkin--Johnsen dynamics. \textbf{a}, Node activities (with the trajectories of the first 5 nodes highlighted by colored lines). \textbf{b}, Adjacent matrix of the network. \textbf{c}, Innate opinions $s_i$ of each node.}
    \label{fig:app_fj}
    \end{center}
    \vskip -0.2in
\end{figure}

Friedkin--Johnsen (FJ) dynamics~\citep{friedkin1990social} is a classic discrete-time opinion formation model on graphs, balancing social influence and innate opinions (stubbornness). The dynamics are defined as:
\begin{equation}
    x_i^{t+1} = \lambda_i\frac{1}{k_i}\sum_{j\in\mathcal{N}(i)}A_{ij}x_j^{t} + (1-\lambda_i)s_i
\end{equation}
where $x_i^t$ is the opinion of node $i$ at time $t$, $k_i=\sum_j A_{ij}$ is the in-degree, and $s_i$ represents the innate opinion. In our setup, $x_i(0), s_i \sim \mathcal{U}(-1, 1)$ and we adopt $\lambda_i = k_i / (k_i + 1)$ with degree-normalized influence. \cref{fig:app_fj} illustrates the opinion evolution and the stubbornness biases present in the network.

\subsubsection{Coupled Map Network (CMN)}
\begin{figure}[ht]
    \vskip 0.2in
    \begin{center}
    \includegraphics[width=0.8\linewidth]{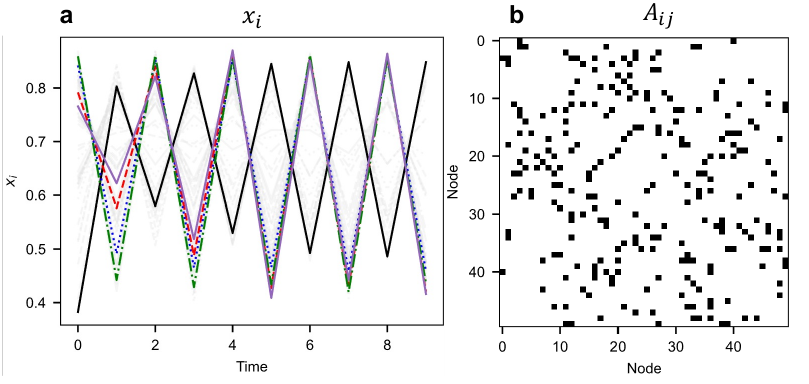}
    \caption{Visual illustration of Coupled map network. \textbf{a}, Node activities (with the trajectories of the first 5 nodes highlighted by colored lines). \textbf{b}, Adjacent matrix of the network.}
    \label{fig:app_cmn}
    \end{center}
    \vskip -0.2in
\end{figure}

Coupled map networks (CMN)~\citep{garcia2002coupled} are discrete-time dynamical systems used to model spatiotemporal chaos and self-organization in complex systems. It couples local chaotic maps through a network structure. The dynamics are defined as:
\begin{equation}
    \theta_{i}^{t+1}= (1-s)f(\theta_{i}^{t}) + \frac{s}{k_{i}}\sum_{j\in\mathcal{N}(i)}A_{ij}f(\theta_{j}^{t})
\end{equation}
where $\theta_i^t$ is the state of node $i$ at time $t$ and $s \in [0,1]$ is the coupling parameter. We use the logistic map $f(x) = \lambda x(1-x)$ as the local nonlinear map with parameters $\lambda = 3.5$ and $s = 0.2$. The system balances local mapping with degree-normalized neighborhood interactions. CMNs are powerful tools for investigating the balance between local chaotic dynamics and global synchronization. \cref{fig:app_cmn} show the dynamic patterns and the interaction matrix for the coupled map network.

\subsection{Baselines}

\paragraph{Granger Causality (GC)}
Granger Causality \cite{granger1969investigating} evaluates the directed functional connectivity between nodes by measuring the predictive gain. Specifically, node $j$ is said to "Granger-cause" node $i$ if the future of $x_i$ can be better predicted using the past of both $x_i$ and $x_j$ than using the past of $x_i$ alone. The GC score is typically computed as the log-ratio of the residual variances: $\mathcal{G}_{j \to i} = \ln (\sigma^2_{\text{restricted}} / \sigma^2_{\text{unrestricted}})$. In our experiments, we implement GC using the \texttt{netrd} library, which fits a linear vector autoregressive (VAR) model to the time series data.

\paragraph{Mutual Information (MI)}
Mutual Information \cite{butte1999mutual} is a non-parametric measure of statistical dependence between two random variables. For two node trajectories $X_i$ and $X_j$, it quantifies the amount of information obtained about $X_i$ through observing $X_j$, defined as: $I(X_i; X_j) = \sum_{x_i, x_j} p(x_i, x_j) \ln \frac{p(x_i, x_j)}{p(x_i)p(x_j)}$. We utilize the \texttt{netrd} implementation, which estimates the probability densities to construct a symmetric adjacency matrix representing pairwise dependencies.

\paragraph{Transfer Entropy (TE)}
Transfer Entropy \cite{schreiber2000measuring} is an information-theoretic measure of directed interaction that quantifies the reduction in uncertainty of the future of $X_i$ given the past of $X_j$, over and above the reduction provided by the past of $X_i$ itself. The formula is given by: $TE_{j \to i} = \sum p(x_{i,t+1}, x_{i,t}, x_{j,t}) \ln \frac{p(x_{i,t+1} | x_{i,t}, x_{j,t})}{p(x_{i,t+1} | x_{i,t})}$. This method is implemented via \texttt{netrd} to capture directed, potentially nonlinear information flows.

\paragraph{Neural Relational Inference (NRI)}
NRI \cite{kipf2018neural} is a variational autoencoder-based framework that jointly infers latent interactions and system dynamics. We employ a GNN-based encoder to infer the edge distribution and an MLP-based decoder for state prediction. The model is trained for 500 epochs with a batch size of 64. We set the number of hidden channels to 512. The optimization uses a dual-learning rate strategy: $lr=0.0005$ for the decoder and $lr_a=0.1$ for the structural inference. The learning rate is decayed by a factor of 0.5 every 200 epochs to ensure convergence.

\paragraph{Graph Dynamics Prior (GDP)}
GDP \cite{pan2024gdp} introduces a generative prior to model the graph distribution, which is optimized alongside the dynamics via variational inference. We configure GDP with 512 hidden channels and a batch size of 64. For spectral modeling, we use a \texttt{power} polynomial filter of order $K=4$. Similar to NRI, we use $lr=0.0005$ for the dynamics parameters and a higher learning rate $lr_a=0.1$ for the distribution estimation to accelerate structural recovery.

\paragraph{Relational Inference via Variational Attention (RIVA)}
RIVA \cite{wu2025riva} is a recent Transformer-based model that interprets attention weights as latent interaction strengths within a variational framework. Although originally designed for weighted graphs, we adapted it to the unweighted setting by binarizing the inferred interactions during evaluation. The hidden dimension is set to 512 with a batch size of 64. The dynamics learning rate is set to $lr=0.00005$, while the structural learning rate is $lr_a=0.1$. This configuration allows the model to capture fine-grained asymmetric dependencies between node trajectories.

\subsection{Evolution of Basis Library}

\begin{figure}[t]
\vskip 0.1in
\begin{center}
\centerline{\includegraphics[width=\linewidth]{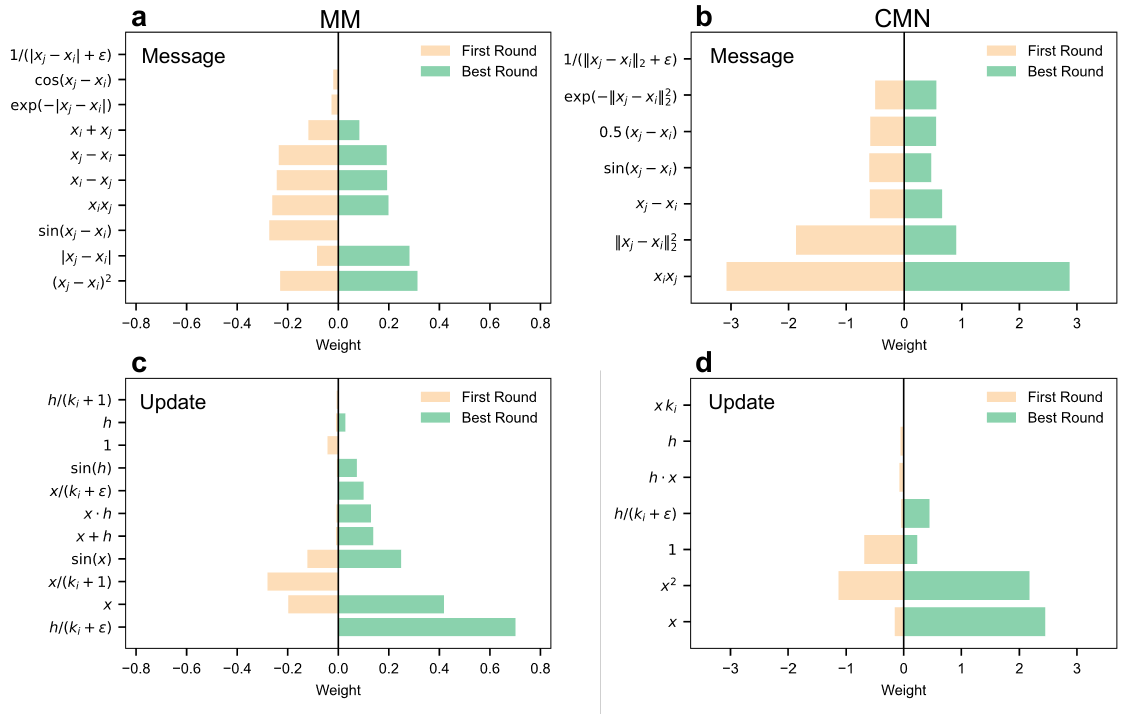}}
\caption{Evolution of discovered basis terms for MM and CMN systems. Panels (a, b) show Message term weights, while (c, d) show Update term weights for the first (blue) and best (red) rounds.}
\label{fig:terms_2}
\end{center}
\vskip -0.1in
\end{figure}

\cref{fig:terms_2} illustrates the evolution of weight distributions within the message and update modules for the \emph{Michaelis--Menten} (MM) and \emph{CMN} systems. While initial weights are dispersed across multiple redundant candidates, the LLM-guided update process drives a sharp focus toward core dynamical primitives in the best-performing round. Specifically, weights concentrate on mechanism-aligned terms—such as degree-normalized reaction inputs in MM ($h/(k_i + \epsilon)$) and accurate coupling or local mapping in CMN ($x_i x_j, x, x^2$)—while significantly suppressing redundant coefficients. This directional convergence demonstrates how the model effectively leverages training feedback to bridge the gap between initial symbolic hypotheses and parsimonious physical representations, which is crucial for capturing complex nonlinear networked dynamics.
Furthermore, \cref{fig:terms_2} presents the distribution of weights in the message and update modules for the \emph{Michaelis--Menten} (MM) and \emph{CMN} systems, comparing the first round of training with the best-performing round. We observe a clear pattern of term evolution and weight rearrangement: in the first round, weights are dispersed across multiple candidate terms, including redundant nonlinearities that are weakly correlated with the true dynamics.

As the LLM-guided update process converges, the weight distributions focus sharply on core dynamical primitives in the best round. Specifically, in the \emph{MM} system, weights focus on terms consistent with degree-normalized reaction inputs (e.g., $h/(k_i + \epsilon)$ in panel c), while in the \emph{CMN} system, the identified terms accurately capture the coupling (e.g., $x_i x_j$ in panel b) and local mapping mechanisms (e.g., $x$ and $x^2$ in panel d). While minor redundant terms may persist, their coefficients are significantly suppressed compared to the dominant, mechanism-aligned terms. This directional convergence demonstrates that the LLM effectively leverages training feedback to bridge the gap between initial symbolic hypotheses and parsimonious physical representations, which is crucial for modeling complex nonlinear networked dynamics.

\subsection{Mechanistic Discovery via Iterative Library Refinement: A Case Study}

\begin{figure}[!t]
    \vskip 0.1in
    \begin{center}
    \includegraphics[width=\linewidth]{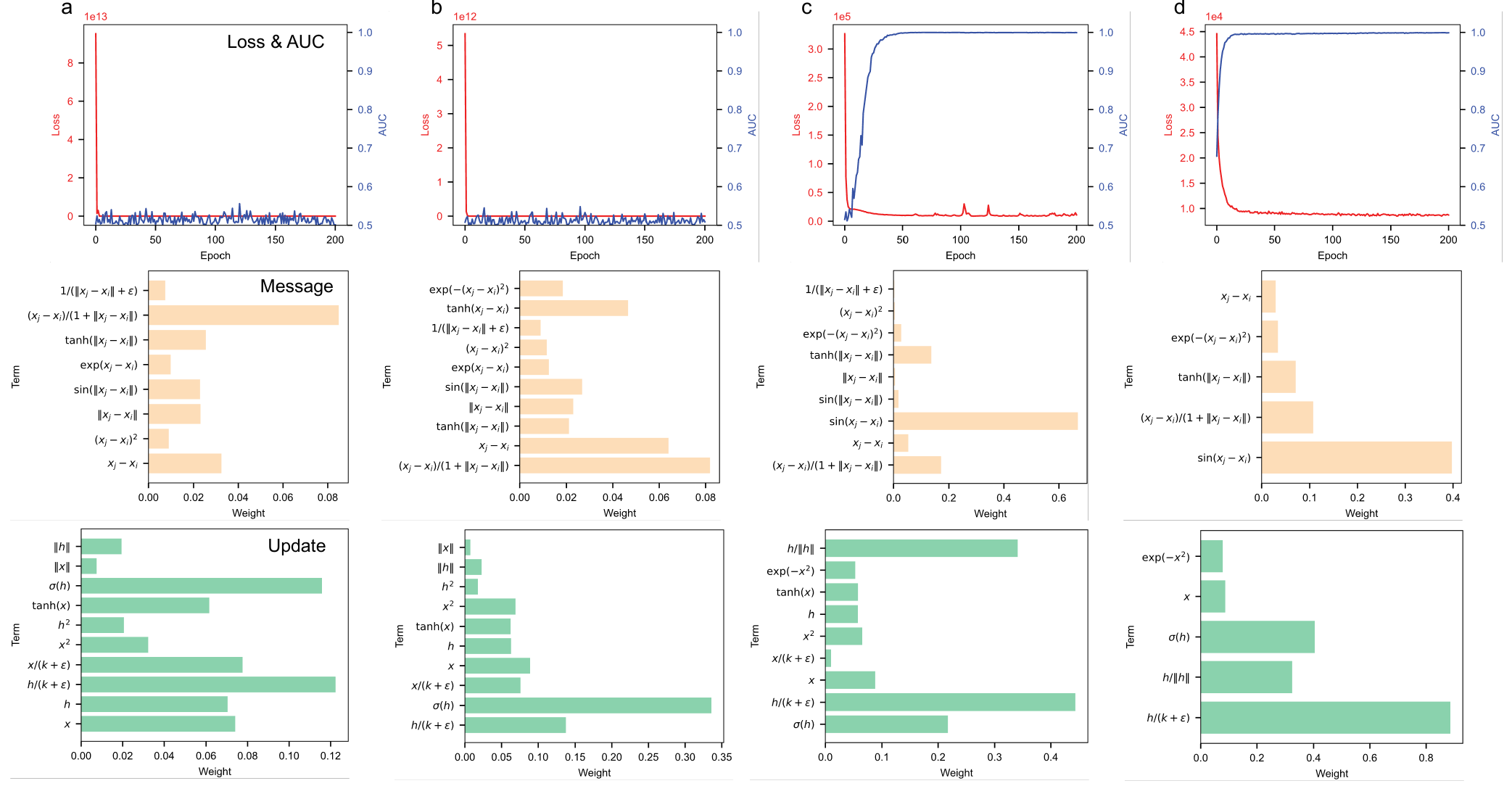}
    \caption{Closed-loop evolution of the basis library in the Kuramoto system in the first 4 rounds (a-d). The first row displays the training loss (red) and structural inference AUC (blue) over four optimization rounds. The second and third rows illustrate the distribution of learned coefficients (importance) for candidate terms in the message and update libraries, respectively. Note the sharp rise in AUC following the introduction of the $\sin(x_j - x_i)$ term and the subsequent pruning of redundant basis functions.}
    \label{fig:kura_evolution}
    \end{center}
    \vskip -0.1in
\end{figure}

To further illustrate the dynamics of the LLM-guided basis library evolution, we track the performance and symbolic composition over four consecutive optimization rounds in the Kuramoto system. As shown in \cref{fig:kura_evolution}, the process demonstrates a clear transition from exploratory hypothesis testing to precise mechanism discovery.

In the initial rounds (Rounds 1 and 2), the message library consists of various candidate terms such as linear differences, tanh, and exponential forms, but lacks the true sinusoidal coupling. During this phase, the structural inference performance (AUC) remains stagnant and suppressed, and the training loss fails to converge to a low steady state. This performance bottleneck confirms that without the correct symbolic primitives, simply increasing the complexity of non-linear fits or structural parameters is insufficient to recover the underlying network.

A significant breakthrough occurs in subsequent rounds when the LLM-guided outer loop introduces the $\sin(x_j - x_i)$ term into the message library. As observed in the training plots, the inclusion of this physically consistent primitive triggers a sharp increase in AUC, rapidly approaching near-perfect recovery (AUC $\approx 1.0$), accompanied by a drastic reduction in prediction loss. In the update module, the model simultaneously identifies that the dynamics are primarily governed by the aggregated messages and degree-normalized terms, effectively capturing the intrinsic natural frequency and coupling effects.

In the final round, the framework performs a pruning operation based on the importance statistics (learned weights) from the previous round. Redundant terms with negligible coefficients are removed, resulting in a highly parsimonious and interpretable library. This evolution highlights the effectiveness of COSINE's closed-loop paradigm: the inner loop provides a quantifiable signal of "how well a symbolic hypothesis works," while the outer loop utilizes this feedback to iteratively refine the expression space, eventually converging to the exact governing laws of the system.

\subsection{Low-Data Performance}
\label{app:low_data}

\begin{table*}[!t]
\caption{Relational inference performance (AUC \%) under low-data regime ($10 \times 10$ samples) across different dynamics and networks.}
\label{tab:low_data_performance}
\vskip 0.1in
\begin{center}
\begin{small}
\begin{sc}
\setlength{\tabcolsep}{6pt}
\renewcommand{\arraystretch}{1.0}
\begin{tabular}{llccccccc}
\toprule
\textbf{Dyn.} & \textbf{Graph} & \textbf{GC} & \textbf{MI} & \textbf{TE} & \textbf{GDP} & \textbf{NRI} & \textbf{RIVA} & \textbf{COSINE} \\
\midrule
\multirow{3}{*}{MM}
 & BA50  & 51.63 & 72.08 & 55.35 & 50.98 & 63.83 & 52.37 & \textbf{72.89} \\
 & ER50  & 54.65 & 64.90 & 53.84 & 52.85 & 59.84 & 52.31 & \textbf{73.21} \\
 & WS50  & 59.48 & \textbf{91.23} & 55.98 & 56.97 & 88.30 & 54.19 & 87.49 \\
\midrule
\multirow{3}{*}{Diff}
 & BA50  & 64.22 & 68.07 & 56.56 & 54.59 & 59.84 & 52.19 & \textbf{91.98} \\
 & ER50  & 58.65 & 56.18 & 53.13 & 63.52 & 81.57 & 52.69 & \textbf{81.72} \\
 & WS50  & 63.66 & 55.68 & 67.76 & 56.16 & 91.17 & 54.97 & \textbf{98.21} \\
\midrule
\multirow{3}{*}{Spr}
 & BA50  & 54.23 & 58.86 & 59.26 & 98.62 & 92.13 & 50.62 & \textbf{99.28} \\
 & ER50  & 50.15 & 50.13 & 52.51 & 98.67 & 78.00 & 51.18 & \textbf{99.39} \\
 & WS50  & 53.81 & 51.40 & 51.50 & 95.18 & 81.53 & 56.64 & \textbf{95.78} \\
\midrule
\multirow{3}{*}{Kura}
 & BA50  & 60.03 & 92.25 & 68.86 & 58.02 & 67.39 & 55.45 & \textbf{96.06} \\
 & ER50  & 56.67 & 86.72 & 68.16 & 53.73 & 84.67 & 52.76 & \textbf{91.10} \\
 & WS50  & 56.30 & 99.59 & 91.09 & 89.82 & 52.48 & 62.63 & \textbf{99.98} \\
\midrule
\multirow{3}{*}{FJ}
 & BA50  & 83.80 & 57.86 & 79.26 & 51.90 & 53.89 & 52.86 & \textbf{94.74} \\
 & ER50  & 87.34 & 54.87 & 75.65 & 53.19 & 51.00 & 50.55 & \textbf{96.26} \\
 & WS50  & 99.00 & 59.73 & 91.02 & 70.76 & 66.27 & 52.37 & \textbf{100.00} \\
\midrule
\multirow{3}{*}{CMN}
 & BA50  & 51.47 & \textbf{86.56} & 71.73 & 78.60 & 76.99 & 53.83 & 60.37 \\
 & ER50  & 54.43 & \textbf{77.47} & 67.91 & 53.99 & 68.56 & 51.04 & 70.09 \\
 & WS50  & 58.63 & \textbf{96.12} & 81.11 & 81.02 & 62.45 & 56.43 & 85.35 \\
\bottomrule
\end{tabular}
\end{sc}
\end{small}
\end{center}
\vskip -0.1in
\end{table*}

\cref{tab:low_data_performance} summarizes relational inference under an extreme low-data regime, where only $10 \times 10$ samples are available for each dynamical system and network. This setting is particularly challenging for approaches that rely on stable estimates of statistical dependence or long-horizon temporal correlations.

Overall, most baselines degrade substantially. Classical statistical and information-theoretic methods (GC, MI, TE) often approach near-chance performance on heterogeneous graphs such as BA50 and ER50, consistent with the difficulty of estimating dependencies from severely limited observations. Model-based and regression-driven approaches (GDP, NRI) are also sensitive to the specific pairing of dynamics and network, performing well in selected cases but lacking uniform robustness.

In contrast, COSINE maintains higher and more stable AUC across all tested systems and networks. The gap is most pronounced for nonlinear and oscillatory dynamics (Diff and Kuramoto), where COSINE achieves near-saturated accuracy even when baselines fluctuate. For mean-field-like dynamics (CMN and MM), where relational signals are intrinsically weaker, COSINE still delivers clear improvements, indicating robustness rather than isolated best-case wins.

These results suggest that COSINE depends less on large sample sizes or long-term statistical convergence. Unlike neural surrogates that require abundant data to constrain model capacity and thus risk overfitting noise in low-data regimes, COSINE imposes strong structural and dynamical inductive biases through explicit symbolic representations, enabling reliable parameter identification from limited observations and improving data efficiency.

From a bias--variance perspective, COSINE deliberately trades expressive flexibility for stronger inductive constraints, which suppress variance when observations are scarce. This bias toward parsimonious, physically meaningful interaction patterns reduces spurious fits and yields more reliable relational inference under severe data scarcity. Although low-data settings are not the primary focus of this work, the consistent gains here further underscore COSINE's robustness and generalization when observations are limited or noisy.

\section{Details of empirical experiment}\label{appd:emp_exp}

\subsection{Discovered Terms}
To elucidate the impact of the LLM-guided hypothesis refinement, we present the initial basis configuration for the COVID-19 dataset in \cref{tab:epi_round000_terms}. By contrasting the first-round candidates with the final converged library, we observe how the LLM-guided agent iteratively prunes redundant terms and augments missing mechanisms. This evolutionary process effectively streamlines the initially cluttered symbolic space, converging toward an optimal and parsimonious basis library that accurately represents the underlying dynamical landscape of the epidemic.

\begin{table*}[!t]
\caption{Discovered terms from round 0 for COVID datasets. Terms are ordered by absolute coefficient magnitude. Each region has Message and Update rows. $x$: node state; $h_i$: aggregated message; $k_i$: degree; $\epsilon=10^{-6}$.}
\label{tab:epi_round000_terms}
\vskip 0.1in
\begin{center}
\begin{small}
\begin{sc}
\setlength{\tabcolsep}{4pt}
\renewcommand{\arraystretch}{1.5}
\resizebox{\textwidth}{!}{%
\begin{tabular}{l c c c c c c c}
\toprule
\textbf{Region} & \textbf{Component} & \textbf{Term 1} & \textbf{Term 2} & \textbf{Term 3} & \textbf{Term 4} & \textbf{Term 5} & \textbf{Term 6} \\
\midrule
\multirow{2}{*}{Arizona}
& Message & $x_i \cdot x_j$ (0.90) & $x_j^{(1)} \cdot (x_j^{(0)} - x_i^{(0)})$ (0.56) & $\|x_j - x_i\|_2$ (0.10) & $\sigma(x_j - x_i)$ (0.05) & $x_j - x_i$ (0.04) & - \\
& Update & $\tanh(x)$ (0.58) & $x \cdot h_i$ (0.13) & $\frac{h_i}{k_i+\epsilon}$ (0.06) & $x^{(2)} \cdot h_i^{(0)}$ (0.05) & $h_i^{(1)}$ (0.03) & - \\
\midrule
\multirow{2}{*}{Connecticut}
& Message & $(x_j - x_i)^2$ (0.54) & $(x_j - x_i)$ (0.10) & $x_j^{(0)}$ (0.07) & $\sigma(\|x_j - x_i\|_2)$ (0.05) & $x_j^{(0)} \cdot \sigma((x_j^{(0)} - x_i^{(0)}))$ (0.04) & - \\
& Update & $x^{(1)}$ (0.23) & $x^{(0)}$ (0.13) & $x^{(2)}$ (0.13) & $x^{(1)} \cdot x^{(2)}$ (0.10) & $\tanh(h_i)$ (0.09) & $\frac{h_i}{k_i+\epsilon}$ (0.07) \\
\midrule
\multirow{2}{*}{Illinois}
& Message & $x_i \cdot x_j$ (0.52) & $\frac{(x_i \cdot (x_j - x_i))}{(1 + \|x_j - x_i\|_2)}$ (0.28) & $\|x_j - x_i\|_2$ (0.07) & $x_j^{(2)} - x_i^{(2)}$ (0.07) & $x_j^{(1)} - x_i^{(1)}$ (0.07) & $2 \cdot (x_j - x_i)$ (0.07) \\
& Update & $-0.1 x$ (0.43) & $-0.2 x^{(1)}$ (0.38) & $\frac{h_i}{k_i+\epsilon}$ (0.10) & $\tanh(h_i)$ (0.09) & $x^{(0)} + 0.5 \cdot x^{(1)}$ (0.08) & $h_i$ (0.04) \\
\midrule
\multirow{2}{*}{Michigan}
& Message & $x_i \cdot (x_j - x_i)$ (0.27) & $x_j^{(1)} \cdot (x_j^{(0)} - x_i^{(0)})$ (0.08) & $(x_j - x_i)^2$ (0.06) & $x_j - x_i$ (0.03) & $\tanh(x_j - x_i)$ (0.01) & - \\
& Update & $-x$ (0.47) & $\frac{h_i}{k_i+\epsilon}$ (0.11) & $x \cdot (1 - x)$ (0.10) & $h_i^{(1)}$ (0.07) & $h_i$ (0.05) & $h_i^{(0)} + h_i^{(1)} + h_i^{(2)}$ (0.03) \\
% \midrule
\bottomrule
\end{tabular}%
}
\end{sc}
\end{small}
\end{center}
\vskip -0.1in
\end{table*}

\subsection{Dataset}

In this section, we provide detailed visualizations and descriptions of the four real-world COVID-19 datasets used in our experiments. Each dataset corresponds to a specific U.S. state, comprising county-level daily confirmed cases and geographic connectivity.

% --- Arizona ---
\subsubsection{Arizona}
\begin{figure}[htbp]
    \centering
    \includegraphics[width=0.95\textwidth]{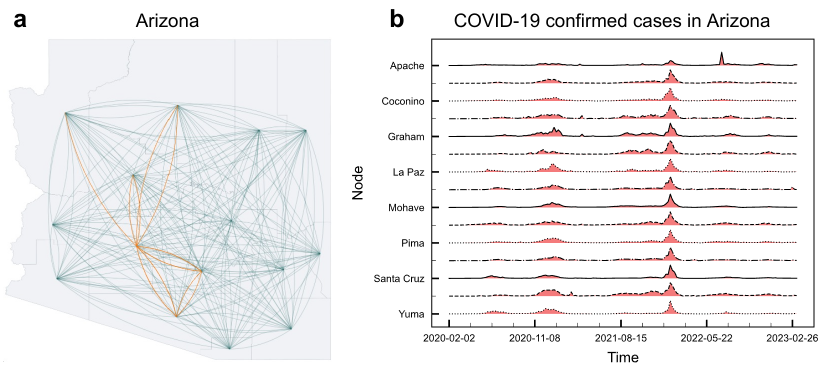}
    \caption{\textbf{Epidemiological data for Arizona.} (a) Geographic map and county-level interaction network. (b) Ridgeline plot of daily COVID-19 confirmed cases across counties from February 2020 to February 2023.}
    \label{fig:appendix_az}
\end{figure}

The Arizona dataset consists of 15 counties. As shown in \cref{fig:appendix_az}a, the geographic network is characterized by large spatial distances between nodes, with a few central hubs (notably Maricopa and Pima counties) exhibiting high connectivity. \cref{fig:appendix_az}b illustrates the temporal distribution of confirmed cases. The state experienced two major synchronized waves: the Winter 2020-2021 wave and the prominent Omicron spike in early 2022. Despite the vast geographic areas, the ridgeline plot reveals a high degree of temporal correlation across counties, suggesting that the primary population centers drive the overall dynamical trend.

% --- Connecticut ---
\subsubsection{Connecticut}
\begin{figure}[htbp]
    \centering
    \includegraphics[width=0.95\textwidth]{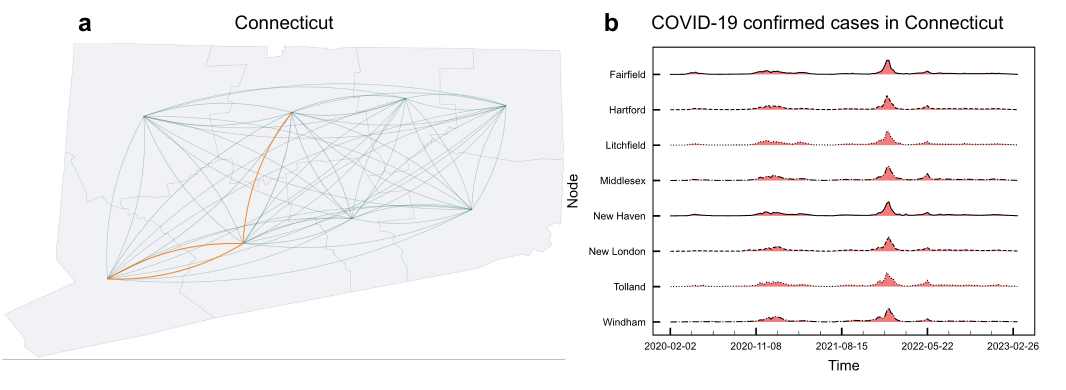}
    \caption{\textbf{Epidemiological data for Connecticut.} (a) Geographic map and county-level interaction network. (b) Ridgeline plot of daily COVID-19 confirmed cases across its 8 counties.}
    \label{fig:appendix_ct}
\end{figure}

Connecticut represents a compact, high-density system with only 8 counties. The network in \cref{fig:appendix_ct}a is relatively dense given its small size, with strong inter-county ties reflecting frequent cross-region commuting. The case distribution in \cref{fig:appendix_ct}b is notable for its extreme synchronization; while earlier waves were relatively subdued, the early 2022 Omicron peak is exceptionally sharp and uniform across all counties (e.g., Fairfield, Hartford, New Haven). This dataset serves as a benchmark for testing relational inference in small-scale, highly coupled systems.

% --- Illinois ---
\subsubsection{Illinois}
\begin{figure}[htbp]
    \centering
    \includegraphics[width=0.95\textwidth]{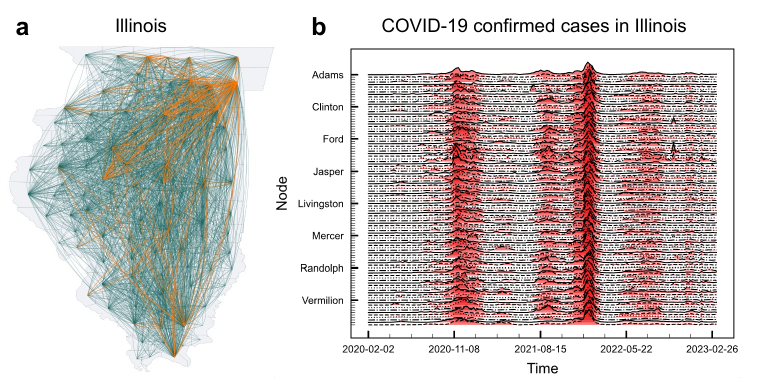}
    \caption{\textbf{Epidemiological data for Illinois.} (a) Geographic map and county-level interaction network. (b) Ridgeline plot showing the high-density case distribution across over 100 counties.}
    \label{fig:appendix_il}
\end{figure}

Illinois provides a large-scale dataset with 102 counties. The interaction network (\cref{fig:appendix_il}a) displays a distinct core-periphery structure, with massive connectivity concentrated in the northeast corner around the Chicago metropolitan area. The temporal data in \cref{fig:appendix_il}b is highly complex, showing multiple distinct regional surges. The large number of nodes (Adams, Clinton, Ford, etc.) creates a dense ridgeline pattern, where the propagation of waves from urban centers to rural peripheries provides a challenging testbed for capturing long-range spatial dependencies.

% --- Michigan ---
\subsubsection{Michigan}
\begin{figure}[!t]
    \centering
    \includegraphics[width=0.95\textwidth]{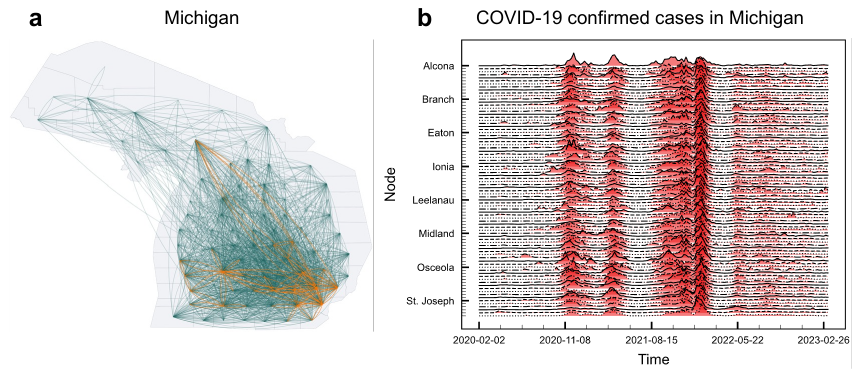}
    \caption{\textbf{Epidemiological data for Michigan.} (a) Geographic map and county-level interaction network. (b) Ridgeline plot of daily COVID-19 confirmed cases illustrating regional seasonal variations.}
    \label{fig:appendix_mi}
\end{figure}

The Michigan dataset includes 83 counties, with its geographic structure constrained by the state's unique peninsula topography. As seen in \cref{fig:appendix_mi}a, the network density is highest in the southeast (the Detroit area), with orange-highlighted edges indicating strong local interaction clusters. \cref{fig:appendix_mi}b shows that Michigan faced unique seasonal patterns, including a significant surge in Spring 2021 that preceded the nationwide Omicron wave. The variation in peak timing and magnitude among counties (e.g., Alcona vs. St. Joseph) allows us to evaluate the model's ability to disentangle local update dynamics from inter-county message passing.

%%%%%%%%%%%%%%%%%%%%%%%%%%%%%%%%%%%%%%%%%%%%%%%%%%%%%%%%%%%%%%%%%%%%%%%%%%%%%%%
%%%%%%%%%%%%%%%%%%%%%%%%%%%%%%%%%%%%%%%%%%%%%%%%%%%%%%%%%%%%%%%%%%%%%%%%%%%%%%%

\end{document}